\documentclass[runningheads]{llncs}

\usepackage{eccv}

\usepackage{eccvabbrv}

\usepackage{graphicx}
\usepackage{booktabs}

\usepackage[accsupp]{axessibility}  

\usepackage{ulem}

\usepackage[symbol]{footmisc}
\DeclareMathOperator*{\argmax}{arg\,max}
\DeclareMathOperator*{\argmin}{arg\,min}
\usepackage{amsfonts}

\usepackage{hyperref}

\usepackage{orcidlink}

\begin{document}

\title{Exploiting Supervised Poison Vulnerability to Strengthen Self-Supervised Defense} 

\titlerunning{VESPR}


\author{Jeremy Styborski\inst{1}$^*$\orcidlink{0009-0005-3465-6213} \and
Mingzhi Lyu\inst{2}$^*$\orcidlink{0009-0005-9359-5908} \and
Yi Huang\inst{1}$^*$\orcidlink{0000-0002-4920-4333} \and 
Adams Kong\inst{1}\orcidlink{0000-0002-9728-9511}}

\authorrunning{J.Styborski et al.}

\institute{College of Computing and Data Science, Nanyang Technological University, Singapore \and
Rapid-Rich Object Search (ROSE) Lab, Interdisciplinary Graduate Programme, Nanyang Technological University, Singapore \\
\email{\{styb0001,lyum0002,AdamsKong\}@ntu.edu.sg, shellbyhuang@gmail.com}}

\maketitle

\def\thefootnote{*}\footnotetext{Equal contribution}
\renewcommand*{\thefootnote}{\arabic{footnote}}

\begin{abstract}
  Availability poisons exploit supervised learning (SL) algorithms by introducing class-related shortcut features in images such that models trained on poisoned data are useless for real-world datasets. Self-supervised learning (SSL), which utilizes augmentations to learn instance discrimination, is regarded as a strong defense against poisoned data. However, by extending the study of SSL across multiple poisons on the CIFAR-10 and ImageNet-100 datasets, we demonstrate that it often performs poorly, far below that of training on clean data. Leveraging the vulnerability of SL to poison attacks, we introduce adversarial training (AT) on SL to obfuscate poison features and guide robust feature learning for SSL. Our proposed defense, designated VESPR (Vulnerability Exploitation of Supervised Poisoning for Robust SSL), surpasses the performance of six previous defenses across seven popular availability poisons. VESPR displays superior performance over all previous defenses, boosting the minimum and average ImageNet-100 test accuracies of poisoned models by 16\% and 9\%, respectively. Through analysis and ablation studies, we elucidate the mechanisms by which VESPR learns robust class features. \textbf{Code:} \ \url{https://github.com/JStyborski/VESPR}
\end{abstract}

\section{Introduction}
\label{sec:Introduction}

The proliferation of open-source labeled data on the Internet has fueled the rapid development and application of deep neural networks across various domains. However, reliance on supervised learning in tasks like image classification exposes neural networks to shortcut learning, whereby models learn `easy' features that are spuriously correlated with task labels in place of useful image features for accurate representation. Shortcut learning is an inherent behavior for neural networks \cite{Tishby_2000_Info_Bottleneck, Geirhos_2018_Texture_Shortcuts, Tian_2020_InfoMin, Lyu_2021_Simplicity_Bias}. Availability poisons exploit this vulnerability by injecting malicious data into real-world data sources, thereby introducing shortcuts and reducing the effectiveness of trained models. Crucially, availability poisons are imperceptible; they only perturb images slightly and do not modify image labels, such that poisoned images appear clean to human observers.

Models trained on poisoned data exhibit poor generalization on real-world clean data, posing significant security risks in critical applications like self-driving vehicles \cite{Patel_2020_Driving_Poison}, anomaly detection \cite{Rubinstein_2009_Anomaly_Poison} and medical AI \cite{Alkhunaizi_2022_Medical_Poison, Adnan_2022_Shortcuts_MI}. For instance, a self-driving system trained with poisoned images of stop signs may fail to recognize real images of stop signs in practical scenarios. A model trained on poisoned data will classify images based on the poisoned patterns rather than genuine features. Consequently, the model becomes incapable of correctly classifying benign images, jeopardizing its reliability in real-world applications. 

Currently, there are various proposed defense methods against availability poisons \cite{DeVries_2017_Cutout, Zhang_2018_Mixup, Yun_2019_Cutmix, Cubuk_2019_RandAugment, Liu_2023_ISS, Qin_2023_UEraser, Kurakin_2017_PGD, Madry_2019_PGD}. However, these methods are limited in their ability to defend against many popular availability poisons, leaving significant vulnerabilities to certain poisons while also suffering from reduced average performance. According to the surveys of poison defenses in Tables \ref{tab:Defense_Literature_CIFAR-10} and \ref{tab:Defense_Literature_ImageNet-100} (see Supplemental Material Sec. \ref{sec:SM_Defense_Literature}), not all of these defense methods have been fully assessed on state-of-the-art poisons. Though most of these baseline defenses achieve robust performance on small-image datasets with few classes (e.g., CIFAR-10 \cite{Krizhevsky_2009_CIFAR} and SVHN \cite{Netzer_2011_SVHN}), our experiments reveal that they often fail to achieve comparable performance on high-resolution datasets with many classes (e.g., ImageNet-100 \cite{Deng_2009_ImageNet}), highlighting a critical research gap.

Poisoned perturbations establish shortcuts between labels and poisonous patterns in the images, prompting us to explore alternatives to traditional SL. SSL methods, such as contrastive learning \cite{Oord_2019_InfoNCE, He_2020_MoCo, Chen_2020_SimCLR}, focus on learning image representations that are invariant to image transformations \cite{Misra_2019_PIRL}, rather than explicitly modeling the correlation between features and labels. In SSL, the combined effect of augmentations and the instance-discrimination objective filter out nuisance features introduced by poisons. Consequently, encoders trained with SSL tend to extract more semantic information from images and avoid class collapse \cite{Xue_2023_Combined_Theory}. In extending the study of SSL-based defense to multiple ImageNet-100 poisons, our experiments reveal that SSL unfortunately learns poison information alongside genuine features, challenging observations made largely on CIFAR-10 \cite{He_2023_Contrastive_Poisoning}.

Motivated by the discrepancy in poison robustness between SL and SSL, we propose leveraging the poison weakness of SL to generate adversarial noise in order to improve augmentations for SSL, thereby preferentially capturing genuine image features and dismissing poisoned features. Thus, we introduce Vulnerability Exploitation of Supervised Poisoning for Robust SSL (VESPR). As illustrated in Fig. \ref{fig:VESPR_Architecture}, VESPR trains an encoder using a multi-task loss function, while adversarial inputs are generated using gradients computed with the supervised cross-entropy loss. We evaluate the performance of VESPR against seven state-of-the-art poisoning methods \cite{Fowl_2021_TAP, Huang_2021_UE, Fu_2022_REM, Yu_2022_LSP, He_2023_Contrastive_Poisoning, Sadasivan_2023_CUDA, Wu_2023_OPS} and compare it with six baseline defense methods \cite{DeVries_2017_Cutout, Zhang_2018_Mixup, Yun_2019_Cutmix, Liu_2023_ISS, Kurakin_2017_PGD, Madry_2019_PGD, He_2023_Contrastive_Poisoning} on the ImageNet-100 \cite{Deng_2009_ImageNet} and CIFAR-10 \cite{Krizhevsky_2009_CIFAR} datasets. Additionally, we conduct comprehensive ablation studies to verify and understand the effectiveness of VESPR.

Our work advances the state-of-the-art for poison defense methods through the following contributions:
\begin{itemize}
    \item We extend the study of multiple defense methods, including SSL, to seven poisons on the CIFAR-10 and ImageNet-100 datasets. For all defenses, we find that the discrepancy between clean and poison training is exacerbated for large-image datasets.
    
    \item We exploit the vulnerability of SL to availability poisons in order to enhance the robustness of SSL training through our proposed method, VESPR. Integrating AT for SL within the SSL framework, VESPR generates optimized adversarial augmentations that preferentially emphasize robust image features and avoid poison shortcuts.
    
    \item Through extensive experiments, we demonstrate that VESPR achieves state-of-the-art performance against multiple prevalent poisons, improving both minimum and average poison defense beyond all other baselines.
\end{itemize}

The remainder of this paper is organized as follows: Sec. \ref{sec:Related_Works} reviews other works relevant to our research, Sec. \ref{sec:Method} analyzes trends in shortcut learning and introduces the VESPR framework, and Sec. \ref{sec:Experiments} covers experimental results, subsequent ablations, and corresponding analyses. 

\section{Related Works}
\label{sec:Related_Works}

\subsection{Data Poisoning}
\label{subsec:RW_Data_Poisoning}

The concept of data poisoning for computer vision was borne out of adversarial examples \cite{Szegedy_2014_PreFGSM, Goodfellow_2015_FGSM, Kurakin_2017_PGD, Madry_2019_PGD}, wherein images could be imperceptibly altered yet decimate supervised classification performance. Importantly, adversarial examples can transfer across models \cite{Demontis_2019_Adversarial_Transfer}, such that generating poisons against unknown training methods becomes feasible. Adversarial Poisons (AP) is an early poisoning method that simply uses adversarial examples \cite{Fowl_2021_TAP}.

Synthetic poisons \cite{Yu_2022_LSP, Sadasivan_2023_CUDA, Wu_2023_OPS} are designed specifically for poisoning supervised classification methods. Synthetic poisons are crafted heuristically and commonly applied class-wise. They aim to introduce consistent class-wise shortcut features. The simplest synthetic poisoning method is the One-Pixel Shortcut \cite{Wu_2023_OPS}, wherein a single pixel is altered to a consistent color for all images of a class. Min-max poisons \cite{Chan-Hon-Tong_2019_RFA, Feng_2019_DeepConfuse, Yuan_2021_NTGA} directly apply the data poisoning objective: find perturbations that minimize training loss on poison data while maximizing test loss on clean data. Min-min poisons \cite{Huang_2021_UE, Fu_2022_REM, He_2023_Contrastive_Poisoning} create `unlearnable examples', such that a network trained on the poisoned samples will quickly learn subtle shortcuts without learning useful image features.

Unlike availability poisons, dirty-label attacks \cite{Carlini_2022_Poisoning_CLIP, Liu_2023_PoisonedEncoder} do not typically modify images but instead inject mislabelled images into datasets. Dirty-label are effective but easily detectable; they can only poison datasets with little or no human supervision. We do not consider dirty-label attacks in this paper and instead focus on poisons that perturb image features.

\subsection{Poison Defenses}
\label{subsec:RW_Poison_Defenses}

Poison defenses for SL are heretofore largely based on augmentations and adversarial training. Most poisons demonstrate reduced effectiveness against models trained with data augmentations like Cutout \cite{DeVries_2017_Cutout}, Mixup \cite{Zhang_2018_Mixup}, CutMix \cite{Yun_2019_Cutmix}, and RandAugment \cite{Cubuk_2019_RandAugment}. Grayscale and JPEG-compression augmentations may be applied to counteract low-frequency and high-frequency poisons, respectively \cite{Liu_2023_ISS}. UEraser samples multiple augmentations and selects the one that maximizes classification loss \cite{Qin_2023_UEraser}. Augmentation-based defenses filter out some non-robust poisonous patterns, but falter against poisons such as AP and CUDA \cite{Fowl_2021_TAP, Sadasivan_2023_CUDA}. In fact, multiple poisoning papers \cite{Fowl_2021_TAP, He_2023_Contrastive_Poisoning, Wu_2023_OPS} demonstrate their triumph over augmentation-based methods as a proof of their effectiveness. Gaussian noise augmentation applied to training data reduces the effectiveness of adversarial examples \cite{Cohen_2019_GaussNoise}. Unlike random noise, AT \cite{Madry_2019_PGD, Uesato_2019_UAT, Yang_2020_Local_Lipschitz} generates optimized adversarial noise during training to improve model robustness. AT is largely considered a state-of-the-art poison defense, improving performance against multiple poisons. However, multiple poison methods still enforce a significant accuracy drop for models defended by AT \cite{He_2023_Contrastive_Poisoning, Sadasivan_2023_CUDA, Wu_2023_OPS}, and some poisons even anticipate the use of AT during training \cite{Fu_2022_REM}. Indeed, we demonstrate in Sec. \ref{subsubsec:E_Architecture Ablation} that these SL-based defenses underperform on multiple poisons.

SSL has also been uncovered as a method for learning robust image features from poisoned data sets \cite{He_2023_Contrastive_Poisoning}. Unfortunately, prior studies are limited in the extent to which they test various poisons, exploring SSL as poison defense against only one ImageNet-100 poison and three CIFAR-10 poisons (see Tables \ref{tab:Defense_Literature_CIFAR-10} and \ref{tab:Defense_Literature_ImageNet-100} in Supplemental Material Sec. \ref{sec:SM_Defense_Literature}). We extend the study of SSL across seven poisons for CIFAR-10 and ImageNet-100. Furthermore, we are the first to investigate adversarial SSL methods \cite{Chen_2020_Pretext_AT, Kim_2020_RoCL, Jiang_2020_ACL, Fan_2021_AdvCL, Addepalli_2022_AT_Augmentation, Luo_2023_DynACL, Kim_2023_TARO} applied to poison defense. 


\subsection{Self-Supervised Learning}
\label{subsec:Self-Supervised_Learning}

Autoencoders \cite{Hinton_2006_AE, Vincent_2008_DAE, Higgins_2017_BVAE, Eastwood_2018_Disentanglement_Eval, Locatello_2019_Inductive_Disentanglement, He_2021_MAE, Kingma_2022_VAE, Yang_2022_ChromaVAE} are considered the one of the first SSL methods, utilizing reconstruction-based loss with an encoder-decoder architecture to learn compressed image embeddings. Recently, invariance-based SSL methods have gained popularity. Contrastive methods \cite{Misra_2019_PIRL, He_2020_MoCo, Chen_2020_SimCLR, Chen_2021_MoCoV3, Radford_2021_CLIP, Tsai_2021_RPC, Tian_2022_AlphaCL, Wang_2022_AlignUnif} typically rely on the InfoNCE loss function \cite{Oord_2019_InfoNCE} to encourage all projections of views from the same source image to be similar, yet different from projections of views from other images. Non-contrastive methods \cite{Grill_2020_BYOL, Chen_2020_SimSiam, Tian_2021_DirectPred, Wang_2022_DirectSet, Zhou_2022_nCLIP, Zhang_2023_Matrix_SSL, Zhuo_2023_Spectral} eliminate the requirement for negative samples, focusing their loss functions entirely on reducing the distance between same-source projections. 



Like SL, SSL is vulnerable to feature suppression \cite{Tian_2020_InfoMin, Li_2022_Collapse, Robinson_2021_Shortcut, Jing_2022_Collapse, Wang_2022_DirectSet, Zhang_2022_NegGrad, Zhuo_2023_Spectral} and shortcut learning \cite{Geirhos_2020_SSL_SL_Texture, Chen_2021_Contrastive_Suppression}. Multiple works have proposed modifications to architecture or training objectives in order to mitigate feature suppression in SSL \cite{Hua_2021_Decorrelation, Kahana_2022_Disentangle_CL, Li_2021_PCL, Robinson_2021_Shortcut, Wang_2021_Disentanglement}. To ensure that SSL does not favor poison shortcuts, we employ AT on SL in order to generate adversarially-perturbed augmentations that shift focus towards robust image features.

\subsection{Multi-Task Learning}
\label{subsec:Multi-Task_Learning}

More recently, multiple groups have proposed combining architectures and loss objectives in order to improve learned representation quality and downstream performance. Multiple works investigate the combination of instance discrimination and reconstruction objectives \cite{Dippel_2022_Contrastive_Recon, Baier_2023_SiDAE, Chen_2024_MixedAE}. Islam et al. \cite{Islam_2021_CL-SL} empirically investigate the performance of supervised losses combined with contrastive SSL. Chen et al. \cite{Chen_2022_Combined_CL} and Xue et al. \cite{Xue_2023_Combined_Theory} both demonstrate that combining supervised and self-supervised contrastive learning objectives avoids class-collapse and mitigates feature suppression. To our knowledge, no one has yet studied multi-task architectures or combined objectives for data poisoning.

\section{Method}
\label{sec:Method}

\subsection{Formulation of Availability Poison}

We describe the availability poison objective in Eq. \ref{eqn:M_Availability_Poison}. We assume a clean dataset $D_c=\{(x_{c,1}, y_1), (x_{c,2}, y_2), ..., (x_{c,N}, y_N)\}$ with distribution $\mathcal{P}_c$, where $x_{c,i}$ is a benign image and $y_i$ is the corresponding class label. The corresponding test dataset is $D'_t=\{(x'_{c,1}, y'_1), (x'_{c,2}, y'_2), ..., (x'_{c,M}, y'_M)\}$, where each pair $(x'_{c,i}, y'_i)$ is drawn from distribution $\mathcal{P}_c$ but not included in $D_c$. The goal of data poisoning is to generate poison dataset $D_p=\{(x_{p,1}, y_1), (x_{p,2}, y_2), ..., (x_{p,N}, y_N)\}$ such that a classification model $CLS_{\theta_{1}^*}(\cdot)$ trained upon poison data $D_p$ will perform poorly on clean test data $D'_t$. Poison sample $x_{p,i}$ is generated from corresponding clean sample $x_{c,i}$ via poisoning process $PSN_{\theta_{2}^*}(x_{c,i})$.
\begin{equation}
    \label{eqn:M_Availability_Poison}
    \begin{split}
  \theta_2^* & = \argmax_{\theta_2} \sum_{i=0}^M L(CLS_{\theta_{1}^*}(x'_{c,i}), y'_i) \\
      \text{s.t.} \quad & \begin{aligned}[t]
      \quad &\theta_1^*  = \argmin_{\theta_1} \sum_{i=0}^N L(CLS_{\theta_{1}}(x_{p,i}),y_i), \\
     &x_{p,i}  = PSN_{\theta_{2}}(x_{c,i}), \quad d(x_{p,i},x_{c,i})  \leq \epsilon \\
  \end{aligned}
  \end{split}
\end{equation}


\noindent
$L$ is the supervised learning loss function. In the context of classification, $L$ is often set as the cross entropy (CE) loss, $L_{CE}(CLS_{\theta_{1}}(x), y) =  y \log CLS_{\theta_{1}}(x) $, where $CLS_{\theta_{1}}(x)$ is the logit of input $x$ with label $y$. Distance metric $d(\cdot)$ between $x_{p,i}$ and $x_{c,i}$ is bounded by a (small) threshold $\epsilon$ such that poisoned images retain high image quality and appear similar to their corresponding clean images. Eq. \ref{eqn:M_Availability_Poison} says that poisoned images contain imperceptible perturbations yet maximally mislead poison-trained models to perform poorly on clean data.

\subsection{Analysis of SL and SSL against Availability Poisons}

\begin{figure}[ht]
\centering
\begin{subfigure}{.45\textwidth}
  \centering
  \includegraphics[width=\linewidth]{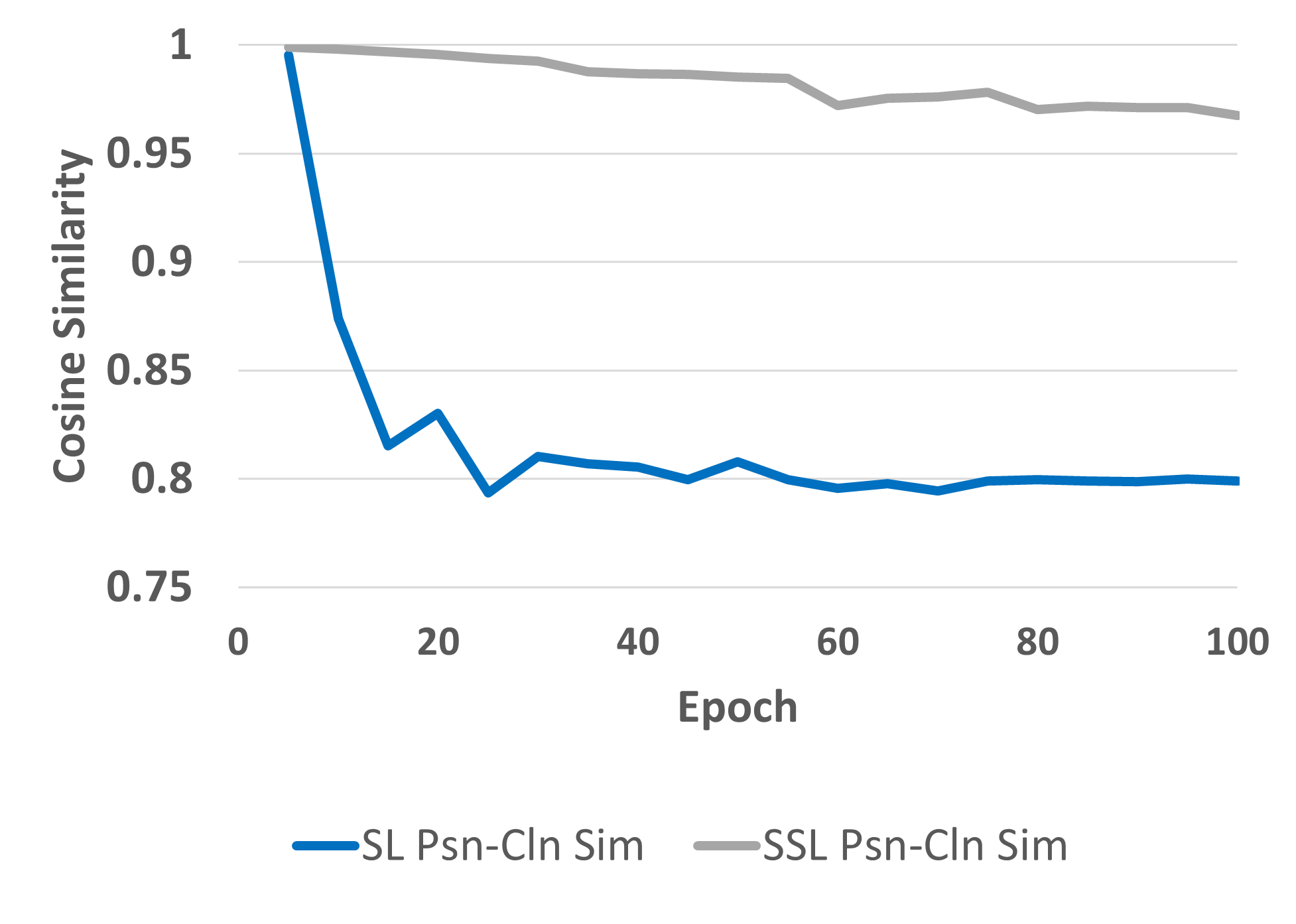}
  \caption{}
  \label{fig:Training_Comparison_A}
\end{subfigure}%
\begin{subfigure}{.45\textwidth}
  \centering
  \includegraphics[width=\linewidth]{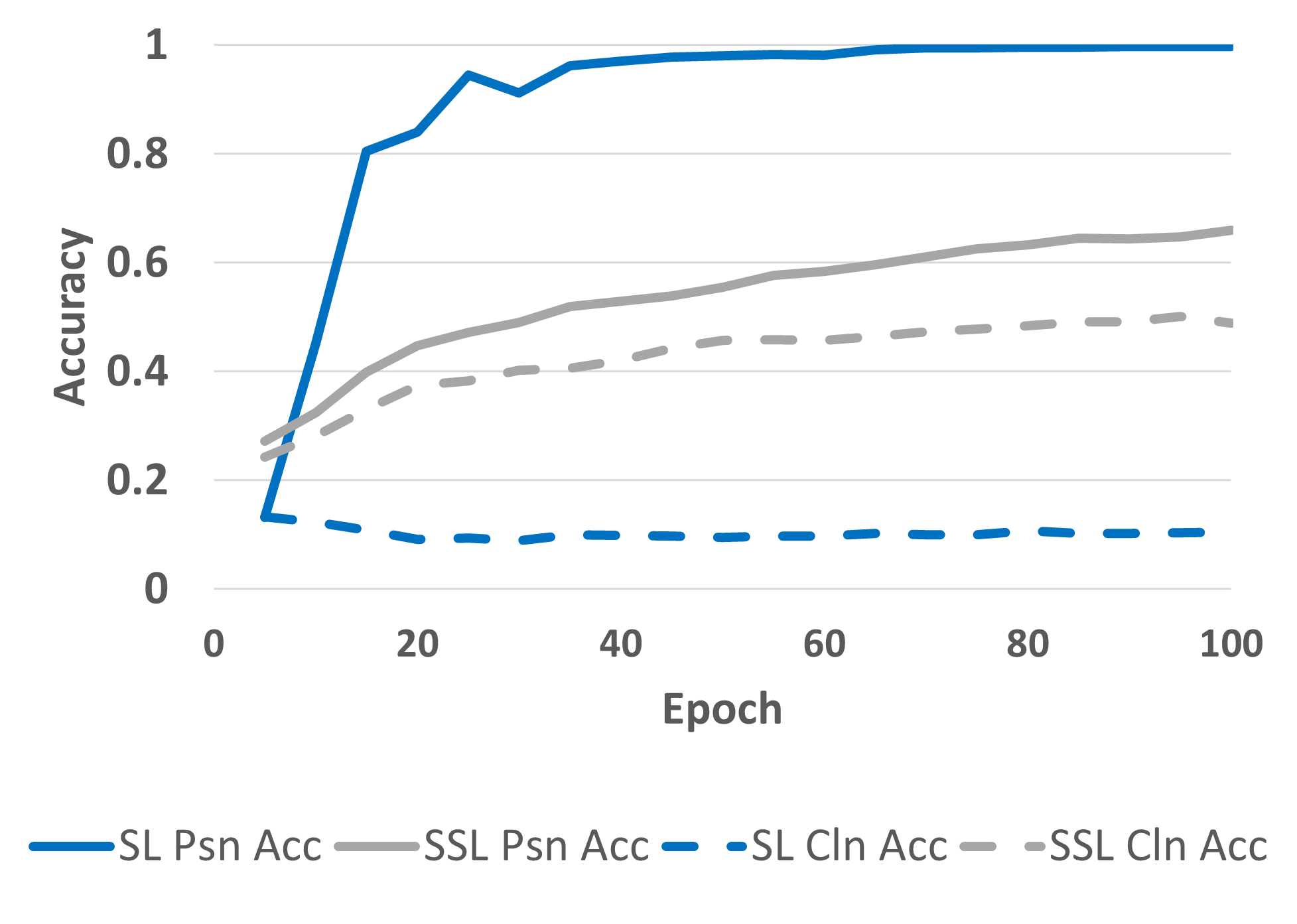}
  \caption{}
  \label{fig:Training_Comparison_B}
\end{subfigure}
\caption{(a) Mean of cosine similarity between representations of clean images and poison images (Psn-Cln Sim, $cossim[f(x_{p,i}),f(x_{c,i})]$). (b) Poison (Psn) and clean (Cln) classification accuracies tested with $x_{p,i}$ and $x_{c,i}$, respectively. We display curves for SL and SSL models, both trained on adversarial poison \cite{Fowl_2021_TAP} data.}
\label{fig:Training_Comparison}
\end{figure}

Intuitively, if there exist features with high task correlation and low variance across training instances, deep learning models will grant greater weight on these features. `Easy' features that are spuriously correlated with the training task become shortcuts, which are quickly learned by SL models. Simplicity bias due to stochastic gradient descent may strengthen the preference for learning shortcuts \cite{Lyu_2021_Simplicity_Bias, Xue_2023_Combined_Theory}. Availability poisons exploit the shortcut-learning weakness in order to mislead model training.

In order to demonstrate the effects of shortcut learning, we train a ResNet18 \cite{He_2015_ResNet} encoder with an adversarial poison \cite{Fowl_2021_TAP} via SL and compare representations of clean and poison images. As shown in Fig. \ref{fig:Training_Comparison_A}, representation similarity between a benign image $x_{c,i}$ and its poisoned version $x_{p,i}$ reduces rapidly (solid blue line in Fig. \ref{fig:Training_Comparison_A}), suggesting that the encoder quickly learns distinct representations based on the shortcut signals in the poisoned images. Though poison accuracy (solid blue line in Fig. \ref{fig:Training_Comparison_B}) quickly maximizes near 100\%, clean accuracy never improves (dashed blue line in Fig. \ref{fig:Training_Comparison_B}).

Since availability poisons are often generated to confuse image classifiers, we explore SSL methods, which bypass the requirement for class labels. Invariance-based SSL techniques are designed to learn representations of images that are robust to input transformations (e.g., cropping, color shifts), yet distinct from representations of other images. 

Contrastive learning is a classic invariance-based SSL framework, backed by extensive research \cite{Misra_2019_PIRL, He_2020_MoCo, Chen_2020_SimCLR, Chen_2021_MoCoV3, Tian_2022_AlphaCL} and proven improvements against some availability poisons \cite{He_2023_Contrastive_Poisoning}. Thus, we primarily utilize contrastive learning for SSL. As noted by Wang and Isola \cite{Wang_2022_AlignUnif}, the contrastive learning objective is optimized by two measures: representation alignment between positive samples and uniformity across all representations. Because SSL is \textbf{not} motivated by class labels, poison features shared by same-class images are less relevant. Additionally, the alignment objective of contrastive learning encourages augmentation invariance; poison features are obfuscated and representation learning favors clean features.

\begin{table*}[h]
  \centering
  \caption{Models are trained on poisoned datasets with SL, SL with SSL-style augmentations, and SSL. Subsequent columns display average cosine similarity between representations of poison images of the same class (Psn In-Cls Sim, $cossim[f(x_{p,i}),f(x_{p,j})]$, where $y_i = y_j$), average cosine similarity between of poison image representations and their corresponding clean image representations (Psn-Cln Sim, $cossim[f(x_{p,i}),f(x_{c,i})]$), and poison representation feature diversity (Psn E-Rank, see \cite{Zhang_2023_Matrix_SSL, Zhuo_2023_Spectral}). We average results across seven different poisons (see Supplemental Material Sec. \ref{subsec:SM_Datasets}).}
  \begin{tabular}{|c|c|c|c|}
    \toprule
    Method & Psn In-Cls Sim & Psn-Cln Sim & Psn E-Rank \\
    \midrule
    SL & 0.78 & 0.70 & 16.67 \\
    SL (SSL Aug) & 0.79 & 0.79 & 16.43 \\
    SSL & 0.47 & 0.86 & 59.61 \\
    \bottomrule
  \end{tabular}
  \label{tab:analysis_SL_SSL}
\end{table*}

In Table \ref{tab:analysis_SL_SSL}, we explore feature suppression in SL and SSL by analyzing representation characteristics for their encoders across seven different poisons. We instantiate the SSL objective $L_{CL}$ with the commonly-used InfoNCE loss \cite{Oord_2019_InfoNCE}. We begin by ablating the effect of augmentation: compared to SL, using SSL-style augmentations during SL barely affects Psn In-Cls Sim and Psn E-Rank, though there is a significant improvement in Psn-Cln Sim. SSL delivers significantly lower Psn In-Cls Sim than SL trained with or without SSL-style augmentation, demonstrating that the SSL training objective (and not SSL-style augmentation) is crucial for encoding instance-specific information rather than learning class-specific shortcuts. Likewise, higher Psn E-Rank and Psn-Cln Sim shows that SSL representations contain more features and are less influenced by poison signals, likely resulting in better generalization on clean test data.

Nevertheless, SSL models cannot inherently differentiate semantic information and poison information, and therefore cannot avoid poison features entirely. This is verified by Fig. \ref{fig:Training_Comparison_B}, where SSL still permits a significant gap between clean and poison accuracies. Therefore, the SSL model has learned some poisoned features along with true image features, leading the downstream classifiers to misclassify data lacking poisonous patterns.

\subsection{VESPR}

Motivated by these observations, we seek methods to improve augmentations for SSL that can better obscure poison features, thereby allowing the learning algorithm to learn true features. Noting the discrepancies between SL and SSL in representation similarities (Table \ref{tab:analysis_SL_SSL}) and performance (Fig. \ref{fig:Training_Comparison_B}), we conclude that the classification network largely learns poison features, distilling them from other image features. This betrays a weakness of availability poisons: once processed by a poisoned encoder, they are no longer subtle nor imperceptible! Drawing from theory for adversarial examples \cite{Goodfellow_2015_FGSM, Fowl_2021_TAP}, we hypothesize that poisoned classifiers can be exploited to generate adversarial examples that maximally muddle poison signals. We further hypothesize that incorporating these adversarial examples as augmentations will improve representations learned by SSL, thereby improving poison defense. Our final architecture is a combined SSL+SL framework, utilizing AT on the supervised loss. We designate our method as VESPR, Vulnerability Exploitation of Supervised Poisons for Robust SSL.

\begin{figure}[t]
  \centering
  \includegraphics[width=\textwidth]{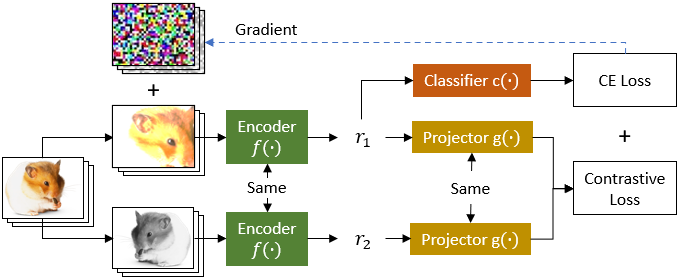}
  \caption{ Overview of VESPR architecture. 
  For each input image, VESPR generates two views through augmentation and encodes them with encoder $f(\cdot)$ as representations $r_1$ and $r_2$, respectively. Cross-entropy (CE) loss is calculated from the output of the classifier head, $c(r_1)$. The CE loss gradient is used to generate adversarial perturbations to the first view using projected gradient descent \cite{Kurakin_2017_PGD, Madry_2019_PGD}. The adversarial images are encoded to replace the original representations. Contrastive loss is calculated from projected representations, $g(r_1)$ and $g(r_2)$, using projections of other images as negative samples. The VESPR network is trained from the combined CE and contrastive losses. 
  }
  \label{fig:VESPR_Architecture}
\end{figure}

As shown in Fig. \ref{fig:VESPR_Architecture}, VESPR contains feature encoder $f(\cdot)$, representation projector $g(\cdot)$, and classification head $c(\cdot)$. During training, each image of a batch is augmented into two different views. One view is taken as the source image to generate adversarial examples using the following objective function:
\begin{equation}
    \label{eqn:M_Adversarial_Perturbation}
    \begin{split}
        \delta^* & = \argmax_\delta L_{CE}(c(f(x+\delta)), y) \\
        & \text{s.t.} \quad   l_p(\delta)  \leq \epsilon_{AT}
    \end{split}
\end{equation}


\noindent
where $\delta$ is an adversarial perturbation, $x$ is an input view, $l_p(\cdot)$ denotes the p-norm, and $\epsilon_{AT}$ is the AT threshold, with $p=\infty$ in our case.

The encoder generates representations of the input views, $f(x+\delta^*)$, $f(x^+)$, and $f(x^-)$. The combined loss function $L_{VESPR}$ is calculated from the outputs of the projector $g(\cdot)$ and classifier $c(\cdot)$ networks:
\begin{equation}
    \label{eqn:M_VESPR}
    \begin{split}
        L_{VESPR}(x+\delta^*, x^+, x^-, y) & = \alpha L_{CL}[g(f(x + \delta^*)), g(f(x^+)), g(f(x^-))] \\
        & + \beta L_{CE}[c(f(x + \delta^*)),y] \\ 
    \end{split}
\end{equation}


\noindent
with inputs as view $x$ with label $y$, corresponding positive view $x^+$ (same source image), and negative views $x^-$ (different source image). $\alpha$ and $\beta$ weight the balance between $L_{CL}$ and $L_{CE}$.

As CE loss is integrated into $L_{VESPR}$, VESPR leverages class-specific features for training. Simultaneously, the contrastive loss component of $L_{VESPR}$ not only focuses on features that distinguish different images, but also mitigates class collapse and poison shortcut learning induced by CE loss \cite{Chen_2022_Combined_CL, Xue_2023_Combined_Theory}. Finally, adversarial samples generated by the classification loss create perturbed augmentations that maximally obscure poison features.

\section{Experiments}
\label{sec:Experiments}

\subsection{Experiment Setup}
\label{subsec:E_Experiment_Setup}

We evaluate the performance of VESPR and other baseline defenses on seven popular poisoning methods: Adversarial Poisons (AP8) \cite{Fowl_2021_TAP}, Contrastive Poisoning (CP8) \cite{He_2023_Contrastive_Poisoning}, Convolution-based Unlearnable Datasets (CUDA) \cite{Sadasivan_2023_CUDA}, Linearly Separable Poisons (LSP16) \cite{Yu_2022_LSP}, One-Pixel Shortcut (OPS) \cite{Wu_2023_OPS}, Robust Unlearnable Examples (RUE8) \cite{Fu_2022_REM}, and Unlearnable Examples (UE8) \cite{Huang_2021_UE}. Further poison descriptions can be found in Supplemental Material Sec. \ref{subsec:SM_Datasets}.

We primarily train models on poisoned versions of a 100-class subset of the ImageNet dataset \cite{Deng_2009_ImageNet}, each with a total of 130,000 training images and 5,000 test images. For benchmarking purposes, we also train models on poisoned CIFAR-10 datasets \cite{Krizhevsky_2009_CIFAR}, each with a total of 50,000 training images and 10,000 test images. Due to the lengthy optimization process required by some poisons on large images, AP8, CP8, and UE8 poisons for ImageNet-100 are generated with 20\% of the dataset, containing 100 classes with a total of 26,000 training images. 

To benchmark the effectiveness of VESPR, we compare against six state-of-the-art poison defense methods: adversarial training (SL+AT), Cutout \cite{DeVries_2017_Cutout}, Mixup \cite{Zhang_2018_Mixup}, CutMix \cite{Yun_2019_Cutmix}, ISS \cite{Liu_2023_ISS}, and SSL (SimCLR) \cite{Chen_2020_SimCLR}. 

For all models, we employ ResNet18 encoders \cite{He_2015_ResNet} with an encoding dimension of 512. When running experiments on CIFAR-10, we follow the architecture of CIFAR ResNet models, as in \cite{Dacosta_2022_SoloLearn}. For all datasets, all SSL methods utilize a 3-layer MLP projection network following the MoCo projector structure \cite{He_2020_MoCo} with hidden and output dimensions of 2048. For classification, we attach a single linear layer to the encoder output. Adversarial examples are generated using projected gradient descent \cite{Kurakin_2017_PGD, Madry_2019_PGD} with 10 steps of size 0.6/255, where perturbations are $\mathit{L_{\infty}}$-bounded at $\epsilon_{AT}=4/255$. Additional training details are deferred to Supplemental Material Sec. \ref{sec:SM_Experiment_Setup}.

We evaluate performance of trained models by measuring classification accuracy on clean test data. For each defense method, we measure the minimum (Psn Min) and average (Psn Avg) accuracies across all poisons. An adversary will choose the most effective poison given state-of-the-art defenses, so increasing Psn Min is crucial to improving model robustness in real-world scenarios.

\subsection{VESPR Performance}
\label{subsec:E_VESPR_Performance}

\begin{table*}[h]
  \centering
  \caption{ImageNet-100 clean test set accuracies for VESPR and six defense benchmarks. Bold numbers indicate the best defense method for each poison, while underlined numbers indicate second-best methods.}
  \begin{tabular}{lcccccccc}
    \toprule
    Poison & SL & SL+AT & Cutout & Mixup & CutMix & ISS & SSL & VESPR \\
    & & & \cite{DeVries_2017_Cutout} & \cite{Zhang_2018_Mixup} & \cite{Yun_2019_Cutmix} & \cite{Liu_2023_ISS} & \cite{Chen_2020_SimCLR} & \\
    \midrule
    Clean & 77.66 & 69.76 & 78.04 & \underline{80.38} & \textbf{81.08} & 71.58 & 70.96 & 74.84 \\
    \midrule
    AP8\cite{Fowl_2021_TAP} & 8.18 & \underline{42.88} & 7.68 & 9.68 & 8.38 & 24.22 & 41.80 & \textbf{48.68} \\
    CP8\cite{He_2023_Contrastive_Poisoning} & 57.46 & 48.76 & 58.20 & \underline{61.76} & \textbf{62.28} & 50.18 & 14.80 & 54.40 \\
    CUDA\cite{Sadasivan_2023_CUDA} & 6.10 & \underline{36.34} & 8.66 & 5.16 & 5.70 & 3.58 & 26.12 & \textbf{45.20} \\
    LSP16\cite{Yu_2022_LSP} & 6.62 & 28.92 & 5.06 & 5.50 & 7.84 & 33.70 & \textbf{62.06} & \underline{56.34} \\
    OPS\cite{Wu_2023_OPS} & 51.00 & 52.56 & 53.86 & 48.82 & \textbf{65.12} & 57.36 & 62.22 & \underline{63.26} \\
    RUE8\cite{Fu_2022_REM} & 14.18 & 57.34 & 15.60 & 33.08 & 16.10 & \underline{67.52} & 65.30 & \textbf{70.36} \\
    UE8\cite{Huang_2021_UE} & 6.32 & 43.26 & 6.42 & 12.38 & 5.80 & 41.18 & \textbf{50.02} & \underline{49.34} \\
    \midrule
    Psn Min & 6.10 & \underline{28.92} & 5.06 & 5.16 & 5.70 & 3.58 & 14.80 & \textbf{45.20} \\
    Psn Avg & 21.41 & 44.29 & 22.21 & 25.20 & 24.46 & 39.68 & \underline{46.05} & \textbf{55.37} \\
    \bottomrule
  \end{tabular}
  \label{tab:VESPR_Performance_ImageNet-100}
\end{table*}

\begin{table*}[h!]
  \centering
  \caption{CIFAR-10 clean test set accuracies for VESPR and six defense benchmarks.}
  \begin{tabular}{lcccccccc}
    \toprule
    Poison & SL & SL+AT & Cutout & Mixup & CutMix & ISS & SSL & VESPR \\
    & & & \cite{DeVries_2017_Cutout} & \cite{Zhang_2018_Mixup} & \cite{Yun_2019_Cutmix} & \cite{Liu_2023_ISS} & \cite{Chen_2020_SimCLR} & \\
    \midrule
    Clean & 93.86 & 89.57 & 95.62 & 95.46 & \textbf{95.78} & 82.40 & 90.55 & 89.35 \\
    \midrule
    AP8\cite{Fowl_2021_TAP} & 12.45 & 85.67 & 9.70 & 33.27 & 8.38 & 81.10 & 75.77 & \textbf{86.78} \\
    CP8\cite{He_2023_Contrastive_Poisoning} & 93.59 & 88.43 & \textbf{94.51} & 93.36 & 93.77 & 82.21 & 73.08 & 89.20 \\
    CUDA\cite{Sadasivan_2023_CUDA} & 21.89 & 48.58 & 23.46 & 21.75 & 24.04 & 21.94 & 66.58 & \textbf{80.31} \\
    LSP16\cite{Yu_2022_LSP} & 21.47 & 85.67 & 18.47 & 22.76 & 21.68 & 79.33 & 88.87 & \textbf{88.96} \\
    OPS\cite{Wu_2023_OPS} & 27.24 & 20.10 & 65.12 & 37.56 & 85.27 & 72.14 & \textbf{87.79} & 84.74 \\
    RUE8\cite{Fu_2022_REM} & 18.91 & 34.45 & 19.67 & 23.27 & 27.46 & 80.04 & 86.70 & \textbf{87.82} \\
    UE8\cite{Huang_2021_UE} & 32.53 & 89.31 & 36.18 & 54.19 & 38.40 & 82.12 & 88.45 & \textbf{89.39} \\
    \midrule
    Psn Min & 18.91 & 20.10 & 9.70 & 21.75 & 8.38 & 21.94 & 66.58 & \textbf{80.31} \\
    Psn Avg & 32.58 & 64.60 & 38.16 & 40.88 & 42.71 & 71.27 & 81.03 & \textbf{86.74} \\
    \bottomrule
  \end{tabular}
  \label{tab:VESPR_Performance_CIFAR-10}
\end{table*}

Table \ref{tab:VESPR_Performance_ImageNet-100} presents clean test accuracy on ImageNet-100 for VESPR as well as multiple baseline defenses. VESPR debuts as a leading defense method, achieving the highest Psn Min and Psn Avg over all previous defenses by 16\% and 9\%, respectively, as well as maintaining strong performance when trained on clean data. VESPR frequently attains the highest clean test accuracy (AP8, CUDA, and RUE) or the second-highest clean test accuracy (LSP16, OPS, and UE8).

Though augmentation-based defenses occasionally achieve top performance (e.g., CutMix on CP8 and OPS), they are often overshadowed by other methods. Indeed the Psn Min values for augmentation-based methods are all less than 10\%. SL+AT and SSL are consistently high performers, achieving average test accuracies of 44.29\% and 46.05\% across all poisons. However, their performance when trained on clean data alone is underwhelming, compared to other methods. This is unfortunate, as it suggests a smaller upper bound for clean test accuracy after poison training; intuitively, a model trained on poison data will not outperform the same model trained on clean data. As expected, SSL is particularly weak to CP8, a poison designed specifically to counter SSL.

Table \ref{tab:VESPR_Performance_CIFAR-10} shows that the trends observed on ImageNet-100 performance are reflected in CIFAR-10. On CIFAR-10, VESPR outperforms other defense methods on multiple poisons (AP8, CUDA, LSP16, RUE8, and UE8). VESPR maintains the highest minimum and average performance across poisons by 13\% and 5\% respectively. For all defenses, the performance gap between clean and poison training is reduced relative to ImageNet-100, likely due to fewer classes and the ease of classifying smaller images.

\subsection{Ablation Studies}
\label{subsec:E_Ablation_Studies}

\subsubsection{Architecture Ablation}
\label{subsubsec:E_Architecture Ablation}

In Table \ref{tab:Architecture_Ablation_ImageNet-100}, we present the poison defense performance of all combinations for SL, SSL, and AT on ImageNet-100. Prior to this work, SSL+AT and SSL+SL methods have not been evaluated for poison defense. As measured by Psn Min and Psn Avg performance, VESPR outperforms all ablations by 13\% and 4\%, respectively. 

\begin{table*}[h]
  \centering
  \caption{ImageNet-100 clean test set accuracies for VESPR and multiple ablations to architecture and training procedure.}
  \begin{tabular}{lccccccc}
    \toprule
    Poison & SL & SL & SL+AT & SSL & SSL+AT & SSL+SL & VESPR \\
    & & (SSL Aug) & & \cite{Chen_2020_SimCLR} & \cite{Jiang_2020_ACL} & & \\
    \midrule
    Clean & 77.66 & 75.48 & 69.76 & 70.96 & 60.86 & \textbf{79.66} & 74.84 \\
    \midrule
    AP8\cite{Fowl_2021_TAP} & 8.18 & 16.54 & 42.88 & 41.80 & 40.46 & 28.82 & \textbf{48.68} \\
    CP8\cite{He_2023_Contrastive_Poisoning} & 57.46 & \textbf{60.40} & 48.76 & 14.80 & 44.00 & 58.38 & 54.40 \\
    CUDA\cite{Sadasivan_2023_CUDA} & 6.10 & 17.30 & 36.34 & 26.12 & 31.78 & 21.82 & \textbf{45.20} \\
    LSP16\cite{Yu_2022_LSP} & 6.62 & 49.30 & 28.92 & \textbf{62.06} & 49.00 & 52.36 & 56.34 \\
    OPS\cite{Wu_2023_OPS} & 51.00 & 63.68 & 52.56 & 62.22 & 53.16 & \textbf{70.14} & 63.26 \\
    RUE8\cite{Fu_2022_REM} & 14.18 & 49.08 & 57.34 & 65.30 & 54.30 & 70.02 & \textbf{70.36} \\
    UE8\cite{Huang_2021_UE} & 6.32 & 34.46 & 43.26 & 50.02 & 40.94 & \textbf{56.68} & 49.34 \\
    \midrule
    Psn Min & 6.10 & 16.54 & 28.92 & 14.80 & 31.78 & 21.82 & \textbf{45.20} \\
    Psn Avg & 21.41 & 41.54 & 44.29 & 46.05 & 44.81 & 51.17 & \textbf{55.37} \\
    \bottomrule
  \end{tabular}
  \label{tab:Architecture_Ablation_ImageNet-100}
\end{table*}

In order to assess the impact of augmentations, we additionally evaluate SL with SSL-style augmentations. For ease of comparison, we compile all augmentation-based defense method results for ImageNet-100 in Table \ref{tab:AugSL_Performance_ImageNet-100} of the Supplemental Material Sec. \ref{subsec:SM_Aug_Defense_Comparison}. As shown in Table \ref{tab:AugSL_Performance_ImageNet-100}, SL with SSL-style augmentations achieves the highest Psn Min and Psn Avg over the other augmentation-based methods by 10\% and 2\%, respectively. However, when compared to VESPR, SSL+SL, SSL, and SL+AT in Table \ref{tab:Architecture_Ablation_ImageNet-100}, SL with SSL-style augmentations consistently underperforms. We hypothesize that existing augmentation-based defenses are weak to data poisoning due to their reliance on image cropping; most poisons are spread throughout images such that all crops are poisoned. This hypothesis is further supported by the fact that all augmentation-based defenses are particularly resistant to OPS, the only poison which does not spread throughout images. This ablation underscores that augmentations cannot fully explain the advances observed with VESPR.

As illustrated in Table \ref{tab:Architecture_Ablation_ImageNet-100}, integrating AT with SL enhances performance over SL alone for all poisons except CP8. Though SL+AT, which trains using bounded additive noise, is particularly effective against bounded additive poisons (e.g., AP8, UE8), improvement is relatively limited on non-additive poisons (e.g. OPS) or poisons with large perturbations (e.g., LSP16). Additionally, applying AT to SSL degrades performance relative to SSL for most poisons. This highlights the nuanced nature of AT application: as poisons aim to undermine supervised classification accuracy, the adversarial signals from SSL alone have limited capability to ignore poisonous patterns.

Aligning with expectations from prior literature, \cite{Chen_2022_Combined_CL, Xue_2023_Combined_Theory}, SSL+SL achieves top performance when trained on clean data, as the combined objectives of SSL and SL prevent the failure modes of either individual objective. Nevertheless, the intuition from prior literature does not necessarily apply for poisoned training, where SSL+SL often performs worse than other defenses, including SSL (e.g., AP8, CUDA, and LSP16). This signifies that the SL component of SSL+SL is still weak to poison patterns, dragging down the performance of SSL+SL. Furthermore, it emphasizes that the additional AT component of VESPR is crucial. For \textbf{all} poisons where SSL+SL underperforms SSL, VESPR utilizes SL-guided adversarial samples to outperform SSL+SL. This validates the claim that poisoned-SL can provide useful guidance for adversarial example generation by obfuscating poison features. 

We defer the corresponding architecture ablation for CIFAR-10 to Table \ref{tab:Architecture_Ablation_CIFAR-10} of the Supplemental Material Sec. \ref{subsec:SM_CIFAR10_Architecture Ablation}. The trends seen for ImageNet-100 in Table \ref{tab:Architecture_Ablation_ImageNet-100} are largely reflected in CIFAR-10, and VESPR attains top performance for Psn Min and Psn Avg across all architecture ablations.

\subsubsection{Architecture Ablation Analysis}

We expand our previous analysis on learned representations (Table \ref{tab:analysis_SL_SSL}) to all architecture ablations in Table \ref{tab:Analysis_All_Avg}. We additionally evaluate representation roughness (i.e., variability) using a local Lipschitz metric (Psn Local Lip) \cite{Yang_2020_Local_Lipschitz}, where lower values correspond to smoother, more robust representation landscapes.

\begin{table*}[h]
  \centering
  \caption{Columns display Psn In-Cls Sim, Psn-Cln Sim, and Psn E-Rank, as in Table \ref{tab:analysis_SL_SSL}. We also present poison image representation roughness (Psn Local Lip, see \cite{Yang_2020_Local_Lipschitz}). We average results across seven different poisons (see Supplemental Material Sec. \ref{subsec:SM_Datasets})}
  \begin{tabular}{|c|c|c|c|c|}
    \toprule
    Method & Psn In-Cls Sim & Psn-Cln Sim & Psn E-Rank & Psn Local Lip \\
    \midrule
    SL & 0.78 & 0.70 & 16.67 & 6180 \\
    SL (SSL Aug) & 0.79 & 0.79 & 16.43 & 5868 \\
    SL+AT & 0.76 & 0.85 & 31.64 & 3312 \\
    SSL & 0.47 & 0.86 & 59.61 & 317 \\
    SSL+AT & 0.60 & 0.97 & 23.47 & 40 \\
    SSL+SL & 0.77 & 0.83 & 34.07 & 8546 \\
    VESPR & 0.72 & 0.91 & 30.29 & 712 \\
    \bottomrule
  \end{tabular}
  \label{tab:Analysis_All_Avg}
\end{table*}

As illustrated in Table \ref{tab:Analysis_All_Avg}, VESPR maintains Psn In-Cls Sim values that are similar to those of SSL+SL and substantially higher than SSL or SSL+AT. Furthermore, utilizing AT consistently increases Psn-Cln Sim and reduces Psn Local Lip across all methods, including VESPR, compared to those without AT. In particular, the Psn Local Lip of VESPR decreases by an order of magnitude, from 8546 to 712, over SSL+SL. This improvement in smoothness is greater than that of any other architecture that incorporates AT, underscoring the superior effect of AT on VESPR. Taken together, the trends in Table \ref{tab:Analysis_All_Avg} suggest that SSL+SL aids VESPR in establishing distinct class-wise clusters of representations while VESPR's SL-guided AT ensures that these representations are reliant on robust, clean features. These findings justify the superior clean-classification performance of VESPR across multiple poisons.

\begin{figure}[h]
\begin{center}
\includegraphics[scale=0.45]{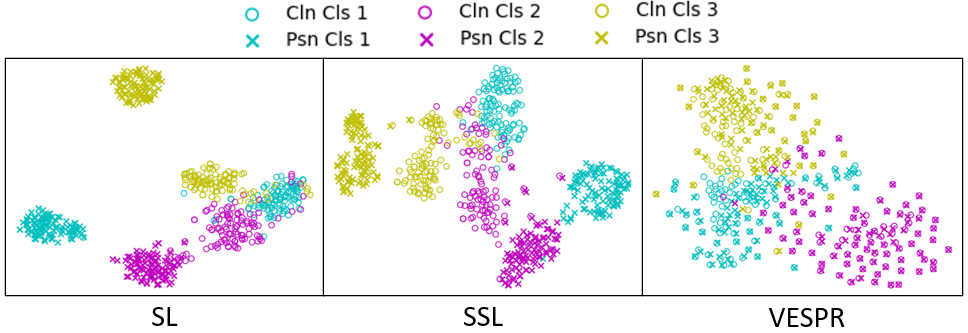}
\end{center}
  \caption{2-dimensional t-SNE plots \cite{Hinton_2002_TSNE} of clean and CUDA-poisoned image representations for SL, SSL, and VESPR models trained on CUDA data.}
  \label{fig:TSNE_Main}
\end{figure}

Trends in representation similarity from Table \ref{tab:Analysis_All_Avg} can also be identified visually. Fig \ref{fig:TSNE_Main} displays 2D t-SNE plots \cite{Hinton_2002_TSNE} of the 512D representations for three classes of poison and clean images and their CUDA-poisoned counterparts for various training methods. Slight mixing between encodings is expected as all three classes are types of fish, yet successful classification requires that clusters are largely distinct. Both SL and SSL tend to tightly cluster poison representations, yet corresponding clean images are encoded to distant representations. Furthermore, all clean image representations from SL and SSL are clustered together, regardless of class, suggesting that clean images are treated as an out-of-distribution class. In contrast, VESPR achieves superior poison-clean alignment, with representations of poison and clean samples overlapping for most samples. This verifies the high Psn-Cln Sim value for VESPR found in Table \ref{tab:Analysis_All_Avg}.

\subsubsection{VESPR Adversarial Ablation}
\label{subsubsec:E_VESPR_Adversarial_Ablation}

In Table \ref{tab:VESPR_Adversarial_Ablation}, we ablate the method of adversarial sample generation. Specifically, we compare the effect of random Gaussian noise (SSL+SL+GN) to methods of adversarial perturbation (VESPR and variants). VESPR remains the top-performing implementation.

For SSL+SL+GN, Gaussian noise with 4/255 standard deviation is applied to images with 50\% probability. As the permissible Gaussian perturbation is higher than the AT perturbation bound, the domain of possible samples generated by Gaussian noise subsumes the domain of samples generated by AT. SSL+SL+GN slightly outperforms SSL+SL on most poisons, but overall performs similarly. Notably, the performance of SSL+SL+GN is conspicuously distinct from that of VESPR, suggesting that gradient-based perturbations from AT are crucial for learning robust features. We note that VESPR consistently outperforms VESPR-SSL and VESPR-Both, with notable improvements on CUDA and RUE8 in particular. VESPR also attains the highest minimum and average clean test accuracies. This suggests that VESPR better exploits the adversarial signal from SL, allowing the encoder to learn robust features with SSL.

\begin{table*}[h]
  \centering
  \caption{ImageNet-100 clean test set accuracy for different methods of sample perturbation. SSL+SL+GN is similar to SSL+SL, but with additional Gaussian noise augmentation. VESPR-SSL uses AT with SSL loss, VESPR-Both uses AT with the combined VESPR loss in Eq. \ref{eqn:M_VESPR}, and VESPR uses AT with SL loss.}
  \begin{tabular}{lccccccc}
    \toprule
    Poison & SSL+SL & SSL+SL+GN & VESPR-SSL & VESPR-Both & VESPR \\
    \midrule
    Clean & \textbf{79.66} & 78.86 & 73.74 & 73.96 & 74.84 \\
    \midrule
    AP8\cite{Fowl_2021_TAP} & 28.82 & 29.92 & 47.56 & 48.02 & \textbf{48.68} \\
    CP8\cite{He_2023_Contrastive_Poisoning} & 58.38 & \textbf{59.14} & 54.20 & 53.64 & 54.40 \\
    CUDA\cite{Sadasivan_2023_CUDA} & 21.82 & 24.02 & 36.64 & 41.26 & \textbf{45.20} \\
    LSP16\cite{Yu_2022_LSP} & 52.36 & 53.00 & 54.36 & 55.88 & \textbf{56.34} \\
    OPS\cite{Wu_2023_OPS} & 70.14 & \textbf{71.12} & 64.30 & 63.14 & 63.26 \\
    RUE8\cite{Fu_2022_REM} & 70.02 & 69.90 & 66.38 & 68.58 & \textbf{70.36} \\
    UE8\cite{Huang_2021_UE} & 56.68 & \textbf{58.80} & 49.28 & 48.20 & 49.34 \\
    \midrule
    Psn Min & 21.82 & 24.02 & 36.64 & 41.26 & \textbf{45.20} \\
    Psn Avg & 51.17 & 52.27 & 53.25 & 54.10 & \textbf{55.37} \\
    \bottomrule
  \end{tabular}
  \label{tab:VESPR_Adversarial_Ablation}
\end{table*}

\section{Conclusion}

We extend the investigation of poison defense to seven poisons on CIFAR-10 and ImageNet-100 datasets, revealing limited performance of current defense methods, including SSL. Drawing from insights on SL and SSL poison defense, we introduce VESPR, which leverages supervised poison vulnerability to enhance self-supervised defense. Through extensive experiments, we demonstrate that VESPR achieves state-of-the-art defense against multiple prevalent poisons, surpassing all other baselines in both minimum and average poison performance.

\clearpage
\section*{Acknowledgements}

This research is supported by the National Research Foundation, Singapore and Infocomm Media Development Authority under its Trust Tech Funding Initiative and Strategic Capability Research Centres Funding Initiative. Any opinions, findings and conclusions or recommendations expressed in this material are those of the author(s) and do not reflect the views of National Research Foundation, Singapore and Infocomm Media Development Authority.
\bibliographystyle{splncs04}
\bibliography{main}

\begin{thebibliography}{10}
\providecommand{\url}[1]{\texttt{#1}}
\providecommand{\urlprefix}{URL }
\providecommand{\doi}[1]{https://doi.org/#1}

\bibitem{Addepalli_2022_AT_Augmentation}
Addepalli, S., Jain, S., Babu, R.V.: Efficient and effective augmentation strategy for adversarial training (2022)

\bibitem{Adnan_2022_Shortcuts_MI}
Adnan, M., Ioannou, Y., Tsai, C.Y., Galloway, A., Tizhoosh, H.R., Taylor, G.W.: Monitoring shortcut learning using mutual information (2022)

\bibitem{Alkhunaizi_2022_Medical_Poison}
Alkhunaizi, N., Kamzolov, D., Tak{\'a}{\v{c}}, M., Nandakumar, K.: Suppressing poisoning attacks on federated learning for medical imaging. In: International Conference on Medical Image Computing and Computer-Assisted Intervention. pp. 673--683. Springer (2022)

\bibitem{Baier_2023_SiDAE}
Baier, F., Mair, S., Fadel, S.G.: Self-supervised siamese autoencoders (2023)

\bibitem{Bardes_2022_VICReg}
Bardes, A., Ponce, J., LeCun, Y.: Vicreg: Variance-invariance-covariance regularization for self-supervised learning (2022)

\bibitem{Carlini_2022_Poisoning_CLIP}
Carlini, N., Terzis, A.: Poisoning and backdooring contrastive learning. In: International Conference on Learning Representations (2022), \url{https://openreview.net/forum?id=iC4UHbQ01Mp}

\bibitem{Caron_2021_DINO}
Caron, M., Touvron, H., Misra, I., Jégou, H., Mairal, J., Bojanowski, P., Joulin, A.: Emerging properties in self-supervised vision transformers (2021)

\bibitem{Chan-Hon-Tong_2019_RFA}
Chan-Hon-Tong, A.: An algorithm for generating invisible data poisoning using adversarial noise that breaks image classification deep learning. Machine Learning and Knowledge Extraction  \textbf{1}(1),  192--204 (2019), \url{https://www.mdpi.com/2504-4990/1/1/11}

\bibitem{Chen_2024_MixedAE}
Chen, K., Liu, Z., Hong, L., Xu, H., Li, Z., Yeung, D.Y.: Mixed autoencoder for self-supervised visual representation learning (2024)

\bibitem{Chen_2022_Combined_CL}
Chen, M.F., Fu, D.Y., Narayan, A., Zhang, M., Song, Z., Fatahalian, K., Ré, C.: Perfectly balanced: Improving transfer and robustness of supervised contrastive learning (2022)

\bibitem{Chen_2020_Pretext_AT}
Chen, T., Liu, S., Chang, S., Cheng, Y., Amini, L., Wang, Z.: Adversarial robustness: From self-supervised pre-training to fine-tuning (2020)

\bibitem{Chen_2020_SimCLR}
Chen, T., Kornblith, S., Norouzi, M., Hinton, G.: A simple framework for contrastive learning of visual representations (2020)

\bibitem{Chen_2021_Contrastive_Suppression}
Chen, T., Luo, C., Li, L.: Intriguing properties of contrastive losses (2021)

\bibitem{Chen_2020_SimSiam}
Chen, X., He, K.: Exploring simple siamese representation learning (2020)

\bibitem{Chen_2021_MoCoV3}
Chen, X., Xie, S., He, K.: An empirical study of training self-supervised vision transformers (2021)

\bibitem{Cohen_2019_GaussNoise}
Cohen, J.M., Rosenfeld, E., Kolter, J.Z.: Certified adversarial robustness via randomized smoothing (2019)

\bibitem{Dacosta_2022_SoloLearn}
da~Costa, V.G.T., Fini, E., Nabi, M., Sebe, N., Ricci, E.: Solo-learn: A library of self-supervised methods for visual representation learning (2022)

\bibitem{Cubuk_2019_RandAugment}
Cubuk, E.D., Zoph, B., Shlens, J., Le, Q.V.: Randaugment: Practical automated data augmentation with a reduced search space (2019)

\bibitem{Demontis_2019_Adversarial_Transfer}
Demontis, A., Melis, M., Pintor, M., Jagielski, M., Biggio, B., Oprea, A., Nita-Rotaru, C., Roli, F.: Why do adversarial attacks transfer? explaining transferability of evasion and poisoning attacks (2019)

\bibitem{Deng_2009_ImageNet}
Deng, J., Dong, W., Socher, R., Li, L.J., Li, K., Fei-Fei, L.: {ImageNet: A Large-Scale Hierarchical Image Database}. In: CVPR09 (2009)

\bibitem{DeVries_2017_Cutout}
DeVries, T., Taylor, G.W.: Improved regularization of convolutional neural networks with cutout (2017)

\bibitem{Dippel_2022_Contrastive_Recon}
Dippel, J., Vogler, S., Höhne, J.: Towards fine-grained visual representations by combining contrastive learning with image reconstruction and attention-weighted pooling (2022)

\bibitem{Eastwood_2018_Disentanglement_Eval}
Eastwood, C., Williams, C.K.I.: A framework for the quantitative evaluation of disentangled representations. In: International Conference on Learning Representations (2018), \url{https://openreview.net/forum?id=By-7dz-AZ}

\bibitem{Fan_2021_AdvCL}
Fan, L., Liu, S., Chen, P.Y., Zhang, G., Gan, C.: When does contrastive learning preserve adversarial robustness from pretraining to finetuning? (2021)

\bibitem{Feng_2019_DeepConfuse}
Feng, J., Cai, Q.Z., Zhou, Z.H.: Learning to confuse: Generating training time adversarial data with auto-encoder (2019)

\bibitem{Fowl_2021_TAP}
Fowl, L., Goldblum, M., yeh Chiang, P., Geiping, J., Czaja, W., Goldstein, T.: Adversarial examples make strong poisons (2021)

\bibitem{Fu_2022_REM}
Fu, S., He, F., Liu, Y., Shen, L., Tao, D.: Robust unlearnable examples: Protecting data privacy against adversarial learning. In: International Conference on Learning Representations (2022), \url{https://openreview.net/forum?id=baUQQPwQiAg}

\bibitem{Geirhos_2020_SSL_SL_Texture}
Geirhos, R., Narayanappa, K., Mitzkus, B., Bethge, M., Wichmann, F.A., Brendel, W.: On the surprising similarities between supervised and self-supervised models (2020)

\bibitem{Geirhos_2018_Texture_Shortcuts}
Geirhos, R., Rubisch, P., Michaelis, C., Bethge, M., Wichmann, F.A., Brendel, W.: Imagenet-trained {CNN}s are biased towards texture; increasing shape bias improves accuracy and robustness. In: International Conference on Learning Representations (2019), \url{https://openreview.net/forum?id=Bygh9j09KX}

\bibitem{Goodfellow_2015_FGSM}
Goodfellow, I.J., Shlens, J., Szegedy, C.: Explaining and harnessing adversarial examples (2015)

\bibitem{Grill_2020_BYOL}
Grill, J.B., Strub, F., Altché, F., Tallec, C., Richemond, P.H., Buchatskaya, E., Doersch, C., Pires, B.A., Guo, Z.D., Azar, M.G., Piot, B., Kavukcuoglu, K., Munos, R., Valko, M.: Bootstrap your own latent: A new approach to self-supervised learning (2020)

\bibitem{He_2023_Contrastive_Poisoning}
He, H., Zha, K., Katabi, D.: Indiscriminate poisoning attacks on unsupervised contrastive learning. In: The Eleventh International Conference on Learning Representations (2023), \url{https://openreview.net/forum?id=f0a_dWEYg-Td}

\bibitem{He_2021_MAE}
He, K., Chen, X., Xie, S., Li, Y., Dollár, P., Girshick, R.: Masked autoencoders are scalable vision learners (2021)

\bibitem{He_2020_MoCo}
He, K., Fan, H., Wu, Y., Xie, S., Girshick, R.: Momentum contrast for unsupervised visual representation learning (2020)

\bibitem{He_2015_ResNet}
He, K., Zhang, X., Ren, S., Sun, J.: Deep residual learning for image recognition (2015)

\bibitem{Higgins_2017_BVAE}
Higgins, I., Matthey, L., Pal, A., Burgess, C., Glorot, X., Botvinick, M., Mohamed, S., Lerchner, A.: beta-{VAE}: Learning basic visual concepts with a constrained variational framework. In: International Conference on Learning Representations (2017), \url{https://openreview.net/forum?id=Sy2fzU9gl}

\bibitem{Hinton_2006_AE}
Hinton, G.E., Salakhutdinov, R.R.: Reducing the dimensionality of data with neural networks. Science  \textbf{313}(5786),  504--507 (2006). \doi{10.1126/science.1127647}, \url{https://www.science.org/doi/abs/10.1126/science.1127647}

\bibitem{Hinton_2002_TSNE}
Hinton, G.E., Roweis, S.: Stochastic neighbor embedding. In: Becker, S., Thrun, S., Obermayer, K. (eds.) Advances in Neural Information Processing Systems. vol.~15. MIT Press (2002), \url{https://proceedings.neurips.cc/paper_files/paper/2002/file/6150ccc6069bea6b5716254057a194ef-Paper.pdf}

\bibitem{Hua_2021_Decorrelation}
Hua, T., Wang, W., Xue, Z., Ren, S., Wang, Y., Zhao, H.: On feature decorrelation in self-supervised learning (2021)

\bibitem{Huang_2021_UE}
Huang, H., Ma, X., Erfani, S.M., Bailey, J., Wang, Y.: Unlearnable examples: Making personal data unexploitable (2021)

\bibitem{Islam_2021_CL-SL}
Islam, A., Chen, C.F., Panda, R., Karlinsky, L., Radke, R., Feris, R.: A broad study on the transferability of visual representations with contrastive learning (2021)

\bibitem{Jiang_2020_ACL}
Jiang, Z., Chen, T., Chen, T., Wang, Z.: Robust pre-training by adversarial contrastive learning (2020)

\bibitem{Jing_2022_Collapse}
Jing, L., Vincent, P., LeCun, Y., Tian, Y.: Understanding dimensional collapse in contrastive self-supervised learning (2022)

\bibitem{Kahana_2022_Disentangle_CL}
Kahana, J., Hoshen, Y.: A contrastive objective for learning disentangled representations (2022)

\bibitem{Kim_2023_TARO}
Kim, M., Ha, H., Son, S., Hwang, S.J.: Effective targeted attacks for adversarial self-supervised learning (2023)

\bibitem{Kim_2020_RoCL}
Kim, M., Tack, J., Hwang, S.J.: Adversarial self-supervised contrastive learning (2020)

\bibitem{Kingma_2022_VAE}
Kingma, D.P., Welling, M.: Auto-encoding variational bayes (2022)

\bibitem{Krizhevsky_2009_CIFAR}
Krizhevsky, A.: Learning multiple layers of features from tiny images (2009), \url{https://api.semanticscholar.org/CorpusID:18268744}

\bibitem{Kurakin_2017_PGD}
Kurakin, A., Goodfellow, I., Bengio, S.: Adversarial examples in the physical world (2017)

\bibitem{Li_2022_Collapse}
Li, A.C., Efros, A.A., Pathak, D.: Understanding collapse in non-contrastive siamese representation learning (2022)

\bibitem{Li_2021_PCL}
Li, T., Fan, L., Yuan, Y., He, H., Tian, Y., Feris, R., Indyk, P., Katabi, D.: Addressing feature suppression in unsupervised visual representations (2021)

\bibitem{Liu_2023_PoisonedEncoder}
Liu, H., Jia, J., Gong, N.Z.: Poisonedencoder: Poisoning the unlabeled pre-training data in contrastive learning (2023), \url{https://arxiv.org/abs/2205.06401}

\bibitem{Liu_2023_ISS}
Liu, Z., Zhao, Z., Larson, M.: Image shortcut squeezing: Countering perturbative availability poisons with compression (2023)

\bibitem{Locatello_2019_Inductive_Disentanglement}
Locatello, F., Bauer, S., Lucic, M., Rätsch, G., Gelly, S., Schölkopf, B., Bachem, O.: Challenging common assumptions in the unsupervised learning of disentangled representations (2019)

\bibitem{Luo_2023_DynACL}
Luo, R., Wang, Y., Wang, Y.: Rethinking the effect of data augmentation in adversarial contrastive learning (2023)

\bibitem{Lyu_2021_Simplicity_Bias}
Lyu, K., Li, Z., Wang, R., Arora, S.: Gradient descent on two-layer nets: Margin maximization and simplicity bias (2021)

\bibitem{Madry_2019_PGD}
Madry, A., Makelov, A., Schmidt, L., Tsipras, D., Vladu, A.: Towards deep learning models resistant to adversarial attacks (2019)

\bibitem{Misra_2019_PIRL}
Misra, I., van~der Maaten, L.: Self-supervised learning of pretext-invariant representations (2019)

\bibitem{Netzer_2011_SVHN}
Netzer, Y., Wang, T., Coates, A., Bissacco, A., Wu, B., Ng, A.Y.: Reading digits in natural images with unsupervised feature learning. In: NIPS Workshop on Deep Learning and Unsupervised Feature Learning 2011 (2011), \url{http://ufldl.stanford.edu/housenumbers/nips2011_housenumbers.pdf}

\bibitem{Oord_2019_InfoNCE}
van~den Oord, A., Li, Y., Vinyals, O.: Representation learning with contrastive predictive coding (2019)

\bibitem{Patel_2020_Driving_Poison}
Patel, N., Krishnamurthy, P., Garg, S., Khorrami, F.: Bait and switch: Online training data poisoning of autonomous driving systems. arXiv preprint arXiv:2011.04065  (2020)

\bibitem{Qin_2023_APBench}
Qin, T., Gao, X., Zhao, J., Ye, K., Xu, C.Z.: Apbench: A unified benchmark for availability poisoning attacks and defenses (2023)

\bibitem{Qin_2023_UEraser}
Qin, T., Gao, X., Zhao, J., Ye, K., Xu, C.Z.: Learning the unlearnable: Adversarial augmentations suppress unlearnable example attacks (2023)

\bibitem{Radford_2021_CLIP}
Radford, A., Kim, J.W., Hallacy, C., Ramesh, A., Goh, G., Agarwal, S., Sastry, G., Askell, A., Mishkin, P., Clark, J., Krueger, G., Sutskever, I.: Learning transferable visual models from natural language supervision (2021)

\bibitem{Robinson_2021_Shortcut}
Robinson, J., Sun, L., Yu, K., Batmanghelich, K., Jegelka, S., Sra, S.: Can contrastive learning avoid shortcut solutions? (2021)

\bibitem{Rubinstein_2009_Anomaly_Poison}
Rubinstein, B.I., Nelson, B., Huang, L., Joseph, A.D., Lau, S.h., Rao, S., Taft, N., Tygar, J.D.: Antidote: understanding and defending against poisoning of anomaly detectors. In: Proceedings of the 9th ACM SIGCOMM Conference on Internet Measurement. pp. 1--14 (2009)

\bibitem{Sadasivan_2023_CUDA}
Sadasivan, V.S., Soltanolkotabi, M., Feizi, S.: Cuda: Convolution-based unlearnable datasets (2023)

\bibitem{Szegedy_2014_PreFGSM}
Szegedy, C., Zaremba, W., Sutskever, I., Bruna, J., Erhan, D., Goodfellow, I., Fergus, R.: Intriguing properties of neural networks (2014)

\bibitem{Tian_2020_InfoMin}
Tian, Y., Sun, C., Poole, B., Krishnan, D., Schmid, C., Isola, P.: What makes for good views for contrastive learning? (2020)

\bibitem{Tian_2022_AlphaCL}
Tian, Y.: Understanding deep contrastive learning via coordinate-wise optimization (2022)

\bibitem{Tian_2021_DirectPred}
Tian, Y., Chen, X., Ganguli, S.: Understanding self-supervised learning dynamics without contrastive pairs (2021)

\bibitem{Tishby_2000_Info_Bottleneck}
Tishby, N., Pereira, F.C., Bialek, W.: The information bottleneck method (2000)

\bibitem{Tsai_2021_RPC}
Tsai, Y.H.H., Ma, M.Q., Yang, M., Zhao, H., Morency, L.P., Salakhutdinov, R.: Self-supervised representation learning with relative predictive coding (2021)

\bibitem{Uesato_2019_UAT}
Uesato, J., Alayrac, J.B., Huang, P.S., Stanforth, R., Fawzi, A., Kohli, P.: Are labels required for improving adversarial robustness? (2019)

\bibitem{Vincent_2008_DAE}
Vincent, P., Larochelle, H., Bengio, Y., Manzagol, P.A.: Extracting and composing robust features with denoising autoencoders. In: Proceedings of the 25th International Conference on Machine Learning. pp. 1096--1103 (01 2008). \doi{10.1145/1390156.1390294}

\bibitem{Wang_2021_Disentanglement}
Wang, T., Yue, Z., Huang, J., Sun, Q., Zhang, H.: Self-supervised learning disentangled group representation as feature (2021)

\bibitem{Wang_2022_AlignUnif}
Wang, T., Isola, P.: Understanding contrastive representation learning through alignment and uniformity on the hypersphere (2022)

\bibitem{Wang_2022_DirectSet}
Wang, X., Chen, X., Du, S.S., Tian, Y.: Towards demystifying representation learning with non-contrastive self-supervision (2022)

\bibitem{Wu_2023_OPS}
Wu, S., Chen, S., Xie, C., Huang, X.: One-pixel shortcut: on the learning preference of deep neural networks (2023)

\bibitem{Xue_2023_Combined_Theory}
Xue, Y., Joshi, S., Gan, E., Chen, P.Y., Mirzasoleiman, B.: Which features are learnt by contrastive learning? on the role of simplicity bias in class collapse and feature suppression (2023)

\bibitem{Yang_2022_ChromaVAE}
Yang, W., Kirichenko, P., Goldblum, M., Wilson, A.G.: Chroma-vae: Mitigating shortcut learning with generative classifiers (2022)

\bibitem{Yang_2020_Local_Lipschitz}
Yang, Y.Y., Rashtchian, C., Zhang, H., Salakhutdinov, R., Chaudhuri, K.: A closer look at accuracy vs. robustness (2020)

\bibitem{Yu_2022_LSP}
Yu, D., Zhang, H., Chen, W., Yin, J., Liu, T.Y.: Availability attacks create shortcuts. In: Proceedings of the 28th {ACM} {SIGKDD} Conference on Knowledge Discovery and Data Mining. {ACM} (aug 2022). \doi{10.1145/3534678.3539241}

\bibitem{Yuan_2021_NTGA}
Yuan, C.H., Wu, S.H.: Neural tangent generalization attacks. In: Meila, M., Zhang, T. (eds.) Proceedings of the 38th International Conference on Machine Learning. Proceedings of Machine Learning Research, vol.~139, pp. 12230--12240. PMLR (18--24 Jul 2021), \url{https://proceedings.mlr.press/v139/yuan21b.html}

\bibitem{Yun_2019_Cutmix}
Yun, S., Han, D., Oh, S.J., Chun, S., Choe, J., Yoo, Y.: Cutmix: Regularization strategy to train strong classifiers with localizable features (2019)

\bibitem{Zhang_2022_NegGrad}
Zhang, C., Zhang, K., Zhang, C., Pham, T.X., Yoo, C.D., Kweon, I.S.: How does simsiam avoid collapse without negative samples? a unified understanding with self-supervised contrastive learning (2022)

\bibitem{Zhang_2018_Mixup}
Zhang, H., Cisse, M., Dauphin, Y.N., Lopez-Paz, D.: mixup: Beyond empirical risk minimization (2018)

\bibitem{Zhang_2023_Matrix_SSL}
Zhang, Y., Tan, Z., Yang, J., Huang, W., Yuan, Y.: Matrix information theory for self-supervised learning (2023)

\bibitem{Zhou_2022_nCLIP}
Zhou, J., Dong, L., Gan, Z., Wang, L., Wei, F.: Non-contrastive learning meets language-image pre-training (2022)

\bibitem{Zhuo_2023_Spectral}
Zhuo, Z., Wang, Y., Ma, J., Wang, Y.: Towards a unified theoretical understanding of non-contrastive learning via rank differential mechanism (2023)

\end{thebibliography}

\clearpage
\appendix{}
\label{sec:SM}

\section{Defense Methods in Literature}
\label{sec:SM_Defense_Literature}

There are significant gaps in literature for defense performance on various poisons. We surveyed multiple works on poison attacks \cite{Fowl_2021_TAP, Huang_2021_UE, Yu_2022_LSP, He_2023_Contrastive_Poisoning, Sadasivan_2023_CUDA, Wu_2023_OPS} and poison defenses \cite{Liu_2023_ISS, Qin_2023_APBench} for data on attack/defense evaluations. Results for CIFAR-10 are displayed in Table \ref{tab:Defense_Literature_CIFAR-10} and results for ImageNet-100 are displayed in Table \ref{tab:Defense_Literature_ImageNet-100}. On CIFAR-10, there is a dearth of data available for the CUDA poison and for SSL as a defense. On ImageNet-100, the lack of data is more severe; CP8, CUDA, and OPS have not been evaluated on most defenses and SSL has not been evaluated across most poisons.

\begin{table*}
  \centering
  \caption{CIFAR-10 poison defense in literature. Individual entries list where data for the corresponding poison and defense method can be found. Empty entries indicate no data found.}
  \begin{tabular}{ccccccccc}
    \toprule
    Poison & SL+AT & Cutout & Mixup & CutMix & ISS & SSL \\
    & & \cite{DeVries_2017_Cutout} & \cite{Zhang_2018_Mixup} & \cite{Yun_2019_Cutmix} & \cite{Liu_2023_ISS} & \\
    \midrule
    AP8\cite{Fowl_2021_TAP} & \cite{Fowl_2021_TAP, Sadasivan_2023_CUDA, Fu_2022_REM, Liu_2023_ISS, Qin_2023_APBench} & \cite{Fowl_2021_TAP, Liu_2023_ISS, Qin_2023_APBench} & \cite{Fowl_2021_TAP, Liu_2023_ISS, Qin_2023_APBench} & \cite{Fowl_2021_TAP, Liu_2023_ISS, Qin_2023_APBench} & \cite{Liu_2023_ISS, Qin_2023_APBench} & \cite{He_2023_Contrastive_Poisoning} \\
    CP8\cite{He_2023_Contrastive_Poisoning} & \cite{He_2023_Contrastive_Poisoning} & \cite{He_2023_Contrastive_Poisoning} & - & - & \cite{Qin_2023_APBench} & \cite{He_2023_Contrastive_Poisoning, Qin_2023_APBench} \\
    CUDA\cite{Sadasivan_2023_CUDA} & \cite{Sadasivan_2023_CUDA} & - & - & - & - & - \\
    LSP16\cite{Yu_2022_LSP} & \cite{Liu_2023_ISS, Qin_2023_APBench} & \cite{Yu_2022_LSP, Liu_2023_ISS, Qin_2023_APBench} & \cite{Yu_2022_LSP, Liu_2023_ISS, Qin_2023_APBench} & \cite{Yu_2022_LSP, Liu_2023_ISS, Qin_2023_APBench} & \cite{Liu_2023_ISS, Qin_2023_APBench} & - \\
    OPS\cite{Wu_2023_OPS} & \cite{Wu_2023_OPS, Liu_2023_ISS, Qin_2023_APBench} & \cite{Wu_2023_OPS, Liu_2023_ISS, Qin_2023_APBench} & \cite{Wu_2023_OPS, Liu_2023_ISS, Qin_2023_APBench} & \cite{Liu_2023_ISS, Qin_2023_APBench} & \cite{Liu_2023_ISS, Qin_2023_APBench} & - \\
    RUE8\cite{Fu_2022_REM} & \cite{Sadasivan_2023_CUDA, Fu_2022_REM, Liu_2023_ISS, Qin_2023_APBench} & \cite{Liu_2023_ISS, Qin_2023_APBench} & \cite{Liu_2023_ISS, Qin_2023_APBench} & \cite{Liu_2023_ISS, Qin_2023_APBench} & \cite{Liu_2023_ISS, Qin_2023_APBench} & - \\
    UE8\cite{Huang_2021_UE} & \cite{Sadasivan_2023_CUDA, Wu_2023_OPS, Fu_2022_REM, Liu_2023_ISS, Qin_2023_APBench} & \cite{Yu_2022_LSP, Wu_2023_OPS, Liu_2023_ISS, Qin_2023_APBench} & \cite{Yu_2022_LSP, Wu_2023_OPS, Liu_2023_ISS, Qin_2023_APBench} & \cite{Yu_2022_LSP, Liu_2023_ISS, Qin_2023_APBench} & \cite{Liu_2023_ISS, Qin_2023_APBench} & \cite{He_2023_Contrastive_Poisoning} \\
    \bottomrule
  \end{tabular}
  \label{tab:Defense_Literature_CIFAR-10}
\end{table*}

\begin{table*}
  \centering
  \caption{ImageNet-100 poison defense in literature. Individual entries list where data for the corresponding poison and defense method can be found. Empty entries indicate no data found.}
  \begin{tabular}{ccccccccc}
    \toprule
    Poison & SL+AT & Cutout & Mixup & CutMix & ISS & SSL \\
    & & \cite{DeVries_2017_Cutout} & \cite{Zhang_2018_Mixup} & \cite{Yun_2019_Cutmix} & \cite{Liu_2023_ISS} & \\
    \midrule
    AP8\cite{Fowl_2021_TAP} & \cite{Sadasivan_2023_CUDA, Fu_2022_REM, Liu_2023_ISS} & \cite{Liu_2023_ISS} & \cite{Liu_2023_ISS} & \cite{Liu_2023_ISS} & \cite{Liu_2023_ISS} & - \\
    CP8\cite{He_2023_Contrastive_Poisoning} & - & - & - & - & - & \cite{He_2023_Contrastive_Poisoning} \\
    CUDA\cite{Sadasivan_2023_CUDA} & \cite{Sadasivan_2023_CUDA} & - & - & - & - & - \\
    LSP16\cite{Yu_2022_LSP} & \cite{Liu_2023_ISS, Qin_2023_APBench} & \cite{Liu_2023_ISS, Qin_2023_APBench} & \cite{Liu_2023_ISS, Qin_2023_APBench} & \cite{Liu_2023_ISS, Qin_2023_APBench} & \cite{Liu_2023_ISS, Qin_2023_APBench} & - \\
    OPS\cite{Wu_2023_OPS} & - & -  & - & - & - & - \\
    RUE8\cite{Fu_2022_REM} & \cite{Sadasivan_2023_CUDA, Fu_2022_REM, Liu_2023_ISS, Qin_2023_APBench} & \cite{Liu_2023_ISS, Qin_2023_APBench} & \cite{Liu_2023_ISS, Qin_2023_APBench} & \cite{Liu_2023_ISS, Qin_2023_APBench} & \cite{Liu_2023_ISS, Qin_2023_APBench} & - \\
    UE8\cite{Huang_2021_UE} & \cite{Sadasivan_2023_CUDA, Fu_2022_REM, Liu_2023_ISS, Qin_2023_APBench} & \cite{Liu_2023_ISS, Qin_2023_APBench} & \cite{Liu_2023_ISS, Qin_2023_APBench} & \cite{Liu_2023_ISS, Qin_2023_APBench} & \cite{Liu_2023_ISS, Qin_2023_APBench} & - \\
    \bottomrule
  \end{tabular}
  \label{tab:Defense_Literature_ImageNet-100}
\end{table*}

\section{Experiment Setup}
\label{sec:SM_Experiment_Setup}

\subsection{Datasets}
\label{subsec:SM_Datasets}

We primarily test our methods on a subset of the ImageNet dataset \cite{Deng_2009_ImageNet}, utilizing the first 100 classes for a total of 130,000 training images and 5,000 test images. We consider ImageNet images to be representative of real-world images. For benchmarking purposes, we additionally measure performance on the CIFAR-10 dataset \cite{Krizhevsky_2009_CIFAR}, with a total of 50,000 training images and 10,000 test images.

We verify against seven popular poisons. When available, we generate poisons using the code presented by corresponding authors; otherwise we attempt to follow the poison formulation as closely as possible. Specifically, we utilize Adversarial Poisons\footnote{https://github.com/lhfowl/adversarial\_poisons} \cite{Fowl_2021_TAP} with 8/255 perturbations, Contrastive Poisoning\footnote{https://github.com/kaiwenzha/contrastive-poisoning} \cite{He_2023_Contrastive_Poisoning} with 8/255 perturbations, Convolution-based Unlearnable Datasets\footnote{https://github.com/vinusankars/Convolution-based-Unlearnability} \cite{Sadasivan_2023_CUDA}, Linearly Separable Poisons\footnote{https://github.com/dayu11/Availability-Attacks-Create-Shortcuts} \cite{Yu_2022_LSP} with 16/255 perturbations, One-Pixel Shortcut\footnote{https://github.com/cychomatica/One-Pixel-Shotcut} \cite{Wu_2023_OPS}, Robust Unlearnable Examples \cite{Fu_2022_REM} with 8/255 perturbations, and Unlearnable Examples\footnote{https://github.com/HanxunH/Unlearnable-Examples} \cite{Huang_2021_UE} with 8/255 perturbations. Due to the lengthy optimization process required to generate some poisons on large images, AP8, CP8, and UE8 poisons for ImageNet-100 are generated as 20\% datasets, containing 100 classes with a total of 26,000 training images. Though RUE8 generation also requires lengthy optimization, the full RUE8 ImageNet-100 dataset is provided by the RUE paper authors \cite{Fu_2022_REM}. 

With the exception of CP8, all poisons studied were designed for SL. The CP8 poison follows the `unlearnable examples' algorithm \cite{Huang_2021_UE} but utilizes SSL architectures and losses \cite{Chen_2020_SimCLR, He_2020_MoCo, Grill_2020_BYOL}. Therefore, the CP8 poison is designed specifically for SSL. Our implementation of CP8 generates sample-wise poisons with SimCLR architecture and loss \cite{Chen_2020_SimCLR}.

\subsection{Implementation}
\label{subsec:SM_Implementation}

For all experiments, we employ a ResNet18 encoder \cite{He_2015_ResNet} with an encoding dimension of 512. When running experiments on CIFAR-10, we follow other CIFAR ResNet models \cite{Dacosta_2022_SoloLearn} by modifying the first convolutional layer to use smaller kernel and stride, and omitting the first maxpool layer. We also implement a momentum encoder for BYOL \cite{Grill_2020_BYOL} with a momentum updates rate of 0.999. All SSL methods utilize a 3-layer MLP projection network following the MoCo projector structure \cite{He_2020_MoCo} with hidden and output dimensions of 2048. For SSL methods that utilize a prediction head (e.g., BYOL, SimSiam), we follow SimSiam \cite{Chen_2020_SimSiam} and use a 2-layer MLP with a hidden dimension of 512 and an output dimension of 2048. For experiments with a classifier, we attach a single linear layer to the encoder output.

\subsection{Loss Formulation}
\label{subsec:SM_Loss_Formulation}

Experiments with BYOL and SimSiam utilize cosine similarity loss with normalized projections. Experiments with SimCLR utilize the InfoNCE loss \cite{Oord_2019_InfoNCE} with normalized projections and a temperature of 0.2. All SSL losses are symmetrized. Classifiers (for linear probe and SL) utilize the cross-entropy loss. SSL+SL combined losses are calculated as in Eq. \ref{eqn:M_VESPR}, with $\alpha=\beta=0.5$.

\subsection{Augmentations}
\label{subsec:SM_Augmentations}

The image augmentation procedure follows SimSiam \cite{Chen_2020_SimSiam}. SSL and SSL+SL pretrain augmentations involve random-resized-crop, color-jitter, grayscale, Gaussian blur, horizontal flip, and normalization. Linear probe augmentations involve random-resized-crop, horizontal flip, and normalization. SL augmentations for ImageNet-100 are identical to those for linear probe, though CIFAR-10 SL augmentations utilize random-crop instead. Test augmentations involve resize, center crop, and normalization. For ImageNet-100, we use a resize value of 256 and a crop size of 224. For CIFAR-10, we maintain the same resize/crop ratio as ImageNet, using a resize value of 32 and a crop size of 28.

\begin{table*}
  \centering
  \caption{Training settings for various methods. For cells with two values, the left value is for CIFAR-10 and the right value is for ImageNet-100.}
  \begin{tabular}{cccccc}
    \toprule
    Setting & SSL+SL & SSL Pretrain & SSL LinProbe & SL \\
    \midrule
    Augmentations & Pretrain & Pretrain & LinProbe & LinProbe \\
    Epochs & 400 | 200 & 400 | 200 & 100 & 100 \\
    Batch Size & 512 | 128 & 512 | 128 & 512 & 256 \\
    LR x BatchSize/256 & 1.0 | 0.25 & 1.0 | 0.25 & 1.0 | 10.0 & 0.1 \\
    LR Warmup & 10 & 10 & 0 & 0 \\
    LR Decay & Cosine & Cosine & Step & Cosine \\
    & & & (0.2x at 60,75,90) & \\
    Optimizer & SGD & SGD & SGD & SGD \\
    Momentum & 0.9 & 0.9 & 0.9 & 0.9 \\
    Weight Decay & 1e-4 & 1e-4 & 0 & 5e-4 \\
    \bottomrule
  \end{tabular}
  \label{tab:Train_Settings}
\end{table*}

\subsection{Training Procedure}
\label{subsec:SM_Training_Procedure}

We present training settings for different training methods in Table \ref{tab:Train_Settings}. We conduct all experiments by first training on poisoned data and then evaluating on clean test data. For SL, SSL+SL, and VESPR methods, we simply evaluate the accuracy of the classifier head. For SSL methods, we evaluate the accuracy of a classifier head finetuned on poisoned data (i.e., linear probe with frozen encoder). Unless otherwise noted, all SSL, SSL+SL, and VESPR methods utilize the SimCLR \cite{Chen_2020_SimCLR} contrastive learning framework. Experiments with SSL largely follow the settings detailed in He et al. \cite{He_2023_Contrastive_Poisoning}. Experiments with SL methods follow the settings given by Liu et al. \cite{Liu_2023_ISS}. 

\subsection{Adversarial Training}
\label{subsec:SM_Adversarial_Training}

Adversarial training largely follows the procedure given in \cite{Liu_2023_ISS}. We use Projected Gradient Descent (PGD) \cite{Kurakin_2017_PGD, Madry_2019_PGD} to generate $\mathit{L_{\infty}}$-bounded perturbations to input augmentations. We employ 10 PGD steps of size 0.6/255, a bound at 4/255, random start, and zero restarts. For AT with SSL, we closely follow the Adversarial Contrastive Learning algorithm \cite{Jiang_2020_ACL}, specifically their A2S method where one augmentation is adversarially perturbed and the other is kept as is. We do not enforce separate batch norms for the adversarial and standard augmentations. Adversarial parameters for SSL are kept identical to those of AT for SL.

\section{Additional Results}
\label{sec:SM_Additional_Results}

\subsection{Augmentation-Based Defense Comparison}
\label{subsec:SM_Aug_Defense_Comparison}

In Table \ref{tab:AugSL_Performance_ImageNet-100}, we concatenate results from Table \ref{tab:VESPR_Performance_ImageNet-100} and Table \ref{tab:Architecture_Ablation_ImageNet-100} to better compare the performance of augmentation-based defenses for SL. SL with SSL augmentations achieves highest minimum and average clean test accuracies across poisons, outperforming other commonly-used defenses.

\begin{table*}
  \centering
  \caption{ImageNet-100 clean test set accuracy across augmentation-based defenses: Cutout, Mixup, CutMix, ISS, and SL with SSL augmentations.}
  \begin{tabular}{ccccccccc}
    \toprule
    Poison & SL & Cutout & Mixup & CutMix & ISS & SL \\
    & & \cite{DeVries_2017_Cutout} & \cite{Zhang_2018_Mixup} & \cite{Yun_2019_Cutmix} & \cite{Liu_2023_ISS} & (SSL Aug) \\
    \midrule
    Clean & 77.66 & 78.04 & 80.38 & \textbf{81.08} & 71.58 & 75.48 \\
    \midrule
    AP8\cite{Fowl_2021_TAP} & 8.18 & 7.68 & 9.68 & 8.38 & \textbf{24.22} & 16.54 \\
    CP8\cite{He_2023_Contrastive_Poisoning} & 57.46 & 58.20 & 61.76 & \textbf{62.28} & 50.18 & 60.40 \\
    CUDA\cite{Sadasivan_2023_CUDA} & 6.10 & 8.66 & 5.16 & 5.70 & 3.58 & \textbf{17.30} \\
    LSP16\cite{Yu_2022_LSP} & 6.62 & 5.06 & 5.50 & 7.84 & 33.70 & \textbf{49.30} \\
    OPS\cite{Wu_2023_OPS} & 51.00 & 53.86 & 48.82 & \textbf{65.12} & 57.36 & 63.68 \\
    RUE8\cite{Fu_2022_REM} & 14.18 & 15.60 & 33.08 & 16.10 & \textbf{67.52} & 49.08 \\
    UE8\cite{Huang_2021_UE} & 6.32 & 6.42 & 12.38 & 5.80 & \textbf{41.18} & 34.46 \\
    \midrule
    Psn Min & 6.10 & 5.06 & 5.16 & 5.70 & 3.58 & \textbf{16.54} \\
    Psn Avg & 21.41 & 22.21 & 25.20 & 24.46 & 39.68 & \textbf{41.54} \\
    \bottomrule
  \end{tabular}
  \label{tab:AugSL_Performance_ImageNet-100}
\end{table*}

\subsection{CIFAR-10 Architecture Ablation}
\label{subsec:SM_CIFAR10_Architecture Ablation}

Table \ref{tab:Architecture_Ablation_CIFAR-10} displays clean test accuracies for VESPR architecture ablations trained on various CIFAR-10 poisons. We find that the trends in poison defense for CIFAR-10 typically mirror those seen for ImageNet-100 in Table \ref{tab:Architecture_Ablation_ImageNet-100}, though with smaller poison-clean discrepancies.

\begin{table*}[h]
  \centering
  \caption{CIFAR-10 clean test set accuracies for VESPR and multiple ablations to architecture and training procedure.}
  \begin{tabular}{cccccccc}
    \toprule
    Poison & SL & SL+AT & SSL & SSL+SL & VESPR \\
    & & & \cite{Chen_2020_SimCLR} & & \\
    \midrule
    Clean & 93.86 & 89.57 & 90.55 & \textbf{94.45} & 89.35 \\
    \midrule
    AP8\cite{Fowl_2021_TAP} & 12.45 & 85.67 & 75.77 & 77.23 & \textbf{86.78} \\
    CP8\cite{He_2023_Contrastive_Poisoning} & \textbf{93.59} & 88.43 & 73.08 & 92.20 & 89.20 \\
    CUDA\cite{Sadasivan_2023_CUDA} & 21.89 & 48.58 & 66.58 & 67.80 & \textbf{80.31} \\
    LSP16\cite{Yu_2022_LSP} & 21.47 & 85.67 & 88.87 & \textbf{90.67} & 88.96 \\
    OPS\cite{Wu_2023_OPS} & 27.24 & 20.10 & 87.79 & \textbf{91.88} & 84.74 \\
    RUE8\cite{Fu_2022_REM} & 18.91 & 34.45 & 86.70 & \textbf{88.47} & 87.82 \\
    UE8\cite{Huang_2021_UE} & 32.53 & 89.31 & 88.45 & \textbf{91.54} & 89.39 \\
    \midrule
    Psn Min & 18.91 & 20.10 & 66.58 & 67.80 & \textbf{80.31} \\
    Psn Avg & 32.58 & 64.60 & 81.03 & 85.68 & \textbf{86.74} \\
    \bottomrule
  \end{tabular}
  \label{tab:Architecture_Ablation_CIFAR-10}
\end{table*}

\subsection{ImageNet-100 Classifier Ablations}

Table \ref{tab:KNN_IN100} quantifies clean test classification performance of VESPR and multiple architecture ablations using a KNN classifier. This is similar to Table \ref{tab:Architecture_Ablation_ImageNet-100} but with a classification method that does \textbf{not} rely on a trained classifier network. To classify a single clean test image, rank the proximity (cosine similarity) of the clean image encoding to 10,240 randomly-sampled poisoned image encodings. The clean image class is decided its 20 nearest neighbors.

Intuitively, KNN accuracy is a distillation of the Psn In-Cls Sim and Psn-Cln Sim metrics from Tables \ref{tab:analysis_SL_SSL} and \ref{tab:Analysis_All_Avg}. If poison image encodings are tightly clustered (high Psn In-Cls Sim) and well-aligned with their clean counterparts (high Psn-Cln Sim), then KNN accuracy should be high. As expected, the KNN accuracy results of Table \ref{tab:KNN_IN100} shadow those of linear classifier accuracy (Table \ref{tab:Architecture_Ablation_ImageNet-100}), and VESPR achieves highest Psn Min and Psn Avg performance. This further demonstrates that VESPR's image encodings are robust to poison features across multiple poisoning methods and supports the analyses from Sec. \ref{subsec:E_Ablation_Studies}. 

\begin{table*}[h]
  \centering
  \caption{ImageNet-100 KNN classifier accuracies for VESPR and multiple ablations. Bold numbers indicate the best defense method for each poison, while underlined numbers indicate second-best methods.}
  \begin{tabular}{lccccccc}
    \toprule
    Poison & SL & SL & SL+AT & SSL & SSL+AT & SSL+SL & VESPR \\
    & & (SSL Aug) & & \cite{Chen_2020_SimCLR} & \cite{Jiang_2020_ACL} & & \\
    \midrule
    Clean & 70.31 & \underline{71.28} & 68.79 & 56.48 & 41.88 & \textbf{78.20} & 69.14 \\
    \midrule
    AP8\cite{Fowl_2021_TAP} & 9.80 & 10.86 & \textbf{47.50} & 35.16 & 24.84 & 28.48 & \underline{44.96} \\
    CP8\cite{He_2023_Contrastive_Poisoning} & 51.33 & \textbf{56.56} & 52.07 & 5.20 & 28.98 & \underline{56.02} & 50.47 \\
    CUDA\cite{Sadasivan_2023_CUDA} & 5.12 & 17.23 & \underline{30.08} & 23.98 & 21.56 & 20.70 & \textbf{41.29} \\
    LSP16\cite{Yu_2022_LSP} & 11.80 & 44.10 & 23.67 & 47.42 & 32.77 & \textbf{52.27} & \underline{49.96} \\
    OPS\cite{Wu_2023_OPS} & 50.12 & \underline{60.70} & 45.43 & 51.17 & 38.01 & \textbf{68.79} & 55.66 \\
    RUE8\cite{Fu_2022_REM} & 13.32 & 41.91 & 53.40 & 53.32 & 40.00 & \textbf{68.16} & \underline{65.90} \\
    UE8\cite{Huang_2021_UE} & 7.27 & 27.07 & \underline{46.64} & 36.80 & 25.63 & \textbf{55.74} & 44.45 \\
    \midrule
    Psn Min & 5.12 & 10.86 & \underline{23.67} & 5.20 & 21.56 & 20.70 & \textbf{41.29} \\
    Psn Avg & 21.25 & 36.92 & 42.68 & 36.15 & 30.26 & \underline{50.02} & \textbf{50.38} \\
    \bottomrule
  \end{tabular}
  \label{tab:KNN_IN100}
\end{table*}

We also ablate the effect of training the VESPR classifier simultaneously with SSL. To this end, we freeze a VESPR-trained encoder and train a new classifier via linear probe on poisoned data with the same settings as SSL linear probe (\ref{tab:Train_Settings}). Table \ref{tab:LinProbe_Ablation} shows that the VESPR encoder with a linear-probe classifier (VESPR+LPC) performs worse than the original VESPR classifier. This is expected; though Tables \ref{tab:Analysis_All_Avg} and \ref{tab:KNN_IN100} demonstrate that VESPR encodings are robust, training a new classifier with poison data allows poison features to determine classifier decision boundaries. Surprisingly, VESPR+LPC also underperforms SSL (which uses poison linear probe). We hypothesize that this is due to higher feature expressivity in SSL (see Psn E-Rank in Table \ref{tab:Analysis_All_Avg}) which may inhibit poison shortcut learning in the classifier. This ablation demonstrates that the VESPR classifier is imperative for poison robustness. Further research on VESPR's feature expressivity is required; possible solutions include uniformity regularization and utilizing other SSL methods (e.g., VICReg \cite{Bardes_2022_VICReg}, DINO \cite{Caron_2021_DINO}) within VESPR.

\begin{table*}[h]
  \centering
  \caption{ImageNet-100 clean test set accuracies for SSL, VESPR, and VESPR encoder with linear probe classifier. Bold values highlight the top-performing methods.}
  \begin{tabular}{lcccccccc}
    \toprule
    Poison & SSL & VESPR & VESPR+LPC \\
    \midrule
    Clean & 70.96 & 74.84 & 69.74 \\
    \midrule
    AP8\cite{Fowl_2021_TAP} & 41.80 & \textbf{48.68} & 44.82 \\
    CP8\cite{He_2023_Contrastive_Poisoning} & 14.80 & \textbf{54.40} & 50.88 \\
    CUDA\cite{Sadasivan_2023_CUDA} & 26.12 & \textbf{45.20} & 20.50 \\
    LSP16\cite{Yu_2022_LSP} & \textbf{62.06} & 56.34 & 48.14 \\
    OPS\cite{Wu_2023_OPS} & 62.22 & \textbf{63.26} & 55.98 \\
    RUE8\cite{Fu_2022_REM} & 65.30 & \textbf{70.36} & 59.80 \\
    UE8\cite{Huang_2021_UE} & \textbf{50.02} & 49.34 & 44.70 \\
    \midrule
    Psn Min & 14.80 & \textbf{45.20} & 20.50 \\
    Psn Avg & 46.05 & \textbf{55.37} & 46.40 \\
    \bottomrule
  \end{tabular}
  \label{tab:LinProbe_Ablation}
\end{table*}

\subsection{SSL Method Ablation}
\label{subsec:SM_SSL_Method_Ablation}

In Table \ref{tab:SSL_Method_Ablation}, we demonstrate the performance of multiple SSL methods. Methods that utilize momentum encoders (MoCo and BYOL) tend to outperform methods that do not (SimCLR and SimSiam). There seems to be no major difference in performance between contrastive methods (SimCLR and MoCo) and non-contrastive methods (SimSiam and BYOL), though contrastive methods tend to give better defense against AP8 and UE8. Relative to pretraining on clean data, all SSL methods perform poorly on multiple poisons, particularly CP8 and CUDA. Unsurprisingly, all SSL methods are weakest against CP8, the poison designed specifically to counter SSL.

Furthermore, we investigate the effect of using different SSL techniques within VESPR. We note that all VESPR variants give similar performance. This is encouraging, as it suggests that the VESPR framework is robust across different SSL techniques. Compared with SSL methods, all VESPR variants provide superior performance on AP8, CP8, and CUDA. Notably, VESPR with BYOL boosts clean test accuracy after CUDA training to 50.14\%, which is the highest CUDA defense value yet seen. All VESPR variants are also resistant to the SSL-based poison, CP8. All VESPR variants boost Psn Min at least 20\% higher than any SSL method. VESPR with SimCLR achieves the highest Psn Avg performance, with an 8\% improvement over the best SSL method (BYOL) and a 0.5\% improvement over the second-best VESPR variant (BYOL).

\begin{table*}[h]
  \centering
  \caption{ImageNet-100 clean test set accuracies for various SSL and VESPR methods. Underlined and bolded values highlight the top-performing SSL and VESPR methods, respectively.}
  \begin{tabular}{lcccccccc}
    \toprule
    Poison & SimCLR & MoCo & SimSiam & BYOL & VESPR & VESPR & VESPR & VESPR \\
    & \cite{Chen_2020_SimCLR} & \cite{Chen_2021_MoCoV3} & \cite{Chen_2020_SimSiam} & \cite{Grill_2020_BYOL} & SimCLR & MoCo & SimSiam & BYOL \\
    \midrule
    Clean & 70.96 & 75.00 & 70.68 & \underline{75.34} & 74.84 & 74.84 & 74.78 & \textbf{75.02} \\
    \midrule
    AP8\cite{Fowl_2021_TAP} & 41.80 & \underline{43.44} & 37.88 & 33.84 & 48.68 & 47.74 & \textbf{48.82} & 48.50 \\
    CP8\cite{He_2023_Contrastive_Poisoning} & 14.80 & 15.80 & 12.82 & \underline{23.64} & 54.40 & 54.20 & 54.26 & \textbf{54.80} \\
    CUDA\cite{Sadasivan_2023_CUDA} & 26.12 & 29.84 & 27.62 & \underline{31.44} & 45.20 & 44.88 & 45.94 & \textbf{50.14} \\
    LSP16\cite{Yu_2022_LSP} & 62.06 & \underline{67.10} & 61.98 & 67.04 & 56.34 & \textbf{56.94} & 51.20 & 50.22 \\
    OPS\cite{Wu_2023_OPS} & 62.22 & 68.04 & 64.54 & \underline{69.00} & \textbf{63.26} & 63.08 & 59.50 & 61.70 \\
    RUE8\cite{Fu_2022_REM} & 65.30 & 70.38 & 66.84 & \underline{70.40} & 70.36 & \textbf{70.56} & 70.04 & 70.34 \\
    UE8\cite{Huang_2021_UE} & 50.02 & \underline{51.04} & 43.92 & 37.64 & \textbf{49.34} & 48.54 & 48.12 & 48.58 \\
    \midrule
    Psn Min & 14.80 & 15.80 & 12.82 & \underline{23.64} & 45.20 & 44.88 & 45.94 & \textbf{48.50} \\
    Psn Avg & 46.05 & \underline{49.38} & 45.09 & 47.57 & \textbf{55.37} & 55.13 & 53.98 & 54.90 \\
    \bottomrule
  \end{tabular}
  \label{tab:SSL_Method_Ablation}
\end{table*}

\subsection{VESPR Loss Weight Ablation}
\label{subsec:SM_Alpha_Ablation}

\begin{figure}[h]
    \centering
    \includegraphics[width=0.75\textwidth]{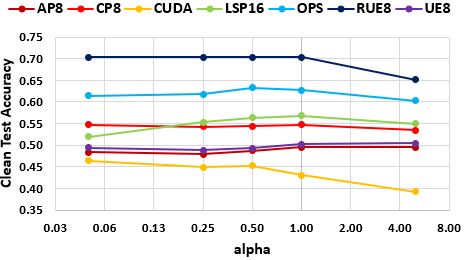}
    \caption{Ablation of the $\alpha$ weight from Eq. \ref{eqn:M_VESPR}. $\beta$ is held constant at $0.5$. The dataset is ImageNet-100.}
    \label{fig:SM_alpha_Ablation}
\end{figure}

Fig. \ref{fig:SM_alpha_Ablation} ablates the value of $\alpha$ from Eq. \ref{eqn:M_VESPR}, effectively varying the relative importance of contrastive loss in VESPR. The set of tested $\alpha$ values is \{0.05, 0.25, 0.50, 1.00, 5.00\}. To ensure that the classifier head is adequately trained, $\beta$ is held constant at $0.5$ throughout the ablations. The dataset is ImageNet-100. Poison robustness is largely consistent for all values of $\alpha$, though it often decreases slightly for excessively large values of $\alpha$. Poison defense is usually highest for moderate $\alpha$ (\{0.25, 0.50, 1.00\}).

\subsection{t-SNE Graphs}
\label{subsec:SM_t-SNE_Graphs}

\begin{figure}
\begin{center}
\includegraphics[scale=0.9]{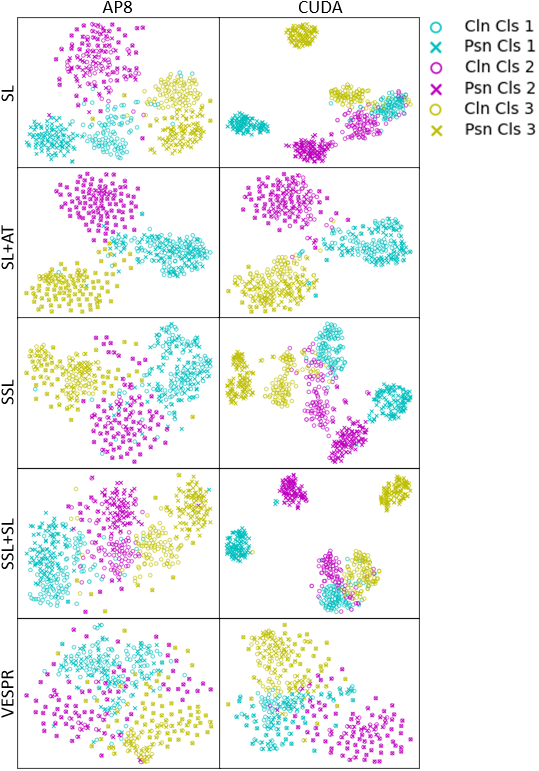}
\end{center}
  \caption{2-dimensional t-SNE plots \cite{Hinton_2002_TSNE} of clean and poison image representations for SL, SL+AT, SSL, SSL+SL, and VESPR models trained on AP8 and CUDA data. Color schemes denote different image classes.}
  \label{fig:TSNE_All}
\end{figure}

Figure \ref{fig:TSNE_All} displays additional t-SNE plots across multiple training methods for both AP8 and CUDA. We note that, across training methods, class clusters for AP8 tend to be more diffuse than those of CUDA, suggesting that CUDA perturbations enforce strong class-based signals. In particular, CUDA enforces strong class separation for SL, SSL, and SSL+SL, with clean representations forming their own out-of-distribution cluster. Alignment between clean representations and poison representations is generally higher for AP8-poisoned encoders. For both poisons, adding AT (e.g., SL+AT or VESPR) greatly improves the alignment between poison and clean representations. VESPR consistently enforces high poison-clean alignment as well as reducing class cluster separation.

\end{document}